\documentclass[a4paper,twoside]{article}

\usepackage{apalike}
\let\bibhang\relax
\usepackage{natbib}
\usepackage{mathtools}
\usepackage{booktabs}
\usepackage{tikz} 

\usepackage{epsfig}
\usepackage{subcaption}
\usepackage{calc}
\usepackage{amssymb}
\usepackage{amstext}
\usepackage{amsmath}
\usepackage{amsthm}
\usepackage{multicol}
\usepackage{pslatex}
\usepackage{algorithm2e}
\usepackage[bottom]{footmisc}
\usepackage{todonotes} 
\usepackage{url}
\usepackage{SCITEPRESS}     

\usepackage{xcolor}

\begin{document}


\title{SACn: Soft Actor-Critic with n-step Returns}


\author{
\authorname{Jakub {\L}yskawa\sup{1}\orcidAuthor{0000-0003-0576-6235},
Jakub Lewandowski\sup{2}, 
Pawe{\l} Wawrzy\'nski\sup{3}\orcidAuthor{0000-0002-1154-0470}
}
\affiliation{\sup{1}Warsaw University of Technology}
\affiliation{\sup{2}Warsaw University}
\affiliation{\sup{3}IDEAS Research Institute}
\email{jakub.lyskawa@pw.edu.pl}
}

\keywords{Reinforcement Learning, Experience Replay, N-Step Return, Importance Sampling}

\abstract{
Soft Actor-Critic (SAC) is widely used in practical applications and is now one of the most relevant off-policy online model-free reinforcement learning (RL) methods. The technique of n-step returns is known to increase the convergence speed of RL algorithms compared to their 1-step returns-based versions. However, SAC is notoriously difficult to combine with n-step returns, since their usual combination introduces bias in off-policy algorithms due to the changes in action distribution. While this problem is solved by importance sampling, a method for estimating expected values of one distribution using samples from another distribution, importance sampling may result in numerical instability. In this work, we combine SAC with n-step returns in a way that overcomes this issue. We present an approach to applying numerically stable importance sampling with simplified hyperparameter selection. Furthermore, we analyze the entropy estimation approach of Soft Actor-Critic in the context of the n-step maximum entropy framework and formulate the $\tau$-sampled entropy estimation to reduce the variance of the learning target. Finally, we formulate the Soft Actor-Critic with n-step returns (SAC$n$) algorithm that we experimentally verify on MuJoCo simulated environments.
}

\onecolumn \maketitle \normalsize \setcounter{footnote}{0} \vfill

\section{\uppercase{Introduction}}\label{sec:intro}

Reinforcement learning (RL) considers the problem of sequential decision making in a dynamic environment. The goal of RL is to
learn to assign the best decision to each environment state. These decisions should account for not only their immediate results but also their further consequences. Although RL is not a new approach, dating back to the methods of dynamic programming \citep{bellman1957dynamic}
, its capabilities are significantly improving through the introduction of new architectures \citep{mnih2015DQN, chen2021transformerRL} and the development of new algorithms \citep{shakya2023algorithmsRL}. As novel RL methods are created and improved, RL is applied to more domains and practical problems, such as large language models \citep{carta2023LLMsRL}, healthcare \citep{jayaraman2024RLinMedicine}, and robotic control \citep{salvato2021RLinRoboticControl}. However,
real-world application of RL still faces numerous challenges, and its feasibility may be improved by introducing more efficient algorithms \citep{dulac-arnold2021challenges}.

In this work, we focus on the online off-policy model-free approach to RL.
Online RL methods improve the decision policy using a trial-and-error approach, interacting with an environment and learning from the experiences, continuously verifying the trained policy. Off-policy algorithms store experience in a buffer and reuse them to increase their efficiency. Model-free algorithms do not directly model the behavior of the environment, and thus, it is not required to design a suitable environment model to apply them to a new environment \citep{sutton2018RL}.

Specifically, in this work, we extend Soft Actor-Critic (SAC) \citep{haarnoja2017sac, haarnoja2017sacAA}, an online off-policy model-free algorithm with an actor-critic architecture that uses the maximum entropy framework to balance exploration and exploitation. SAC is commonly used in many practical applications \citep{
hersi2024multirobotSAC, liang2024SACbasedEMS}, making improvements to this algorithm relevant. In this work, we aim to extend the Soft Actor-Critic by combining it with the n-step returns, an approach that estimates future returns using sequences of rewards from $n$ consecutive time steps.

The main contributions of this paper may be summarized as follows:
\begin{enumerate}
    \item We present the n-step maximum entropy framework, expanding the soft action-value formulation of \citet{haarnoja2017sac}.
    \item We introduce $\tau$-sampled entropy estimation as a method to decrease the variance of Monte Carlo entropy estimation.
    \item We propose a reparameterization approach for the clipping of the importance sampling weights by using a quantile of these weights within a batch to simplify hyperparameter tuning.
    \item We combine the contributions 1-3 into Soft Actor-Critic with n-step Returns (SAC$n$). We experimentally verify SAC$n$ on simulated robotic environments and show that it is robust to hyperparameter values and yields similar or better results than SAC.
\end{enumerate}

\section{\uppercase{Background}}

\subsection{Off-policy reinforcement learning}

In this work, we consider the standard online off-policy reinforcement learning setup with continuous actions. In each time step $t$ the environment is in a state, $s_t$. An agent selects an action, $a_t$, using a policy, $\pi(\cdot|s_t)$, where $\pi(a|s)$ denotes the value of the probability density of the action $a$ in the state $s$. Based on the state $s_t$ and action $a_t$, the environment changes its state in the next time step to $s_{t+1}$ and the agent receives a reward, $r_t \in \mathbb{R}$. Each tuple $\langle s_t, a_t, r_t, s_{t+1}, terminal(s_{t+1})\rangle$ is stored in a replay buffer. If the state $s_{t+1}$ is terminal, which the predicate $terminal(s_{t+1})$ indicates, then the state of the environment is reset.  An episode is a sequence of states, actions, and rewards from the initial environment state to the terminal state.

A sequence of consecutive experience tuples is called a trajectory. In each training step, an online off-policy reinforcement learning algorithm retrieves a batch of trajectories from the replay buffer and uses them to adjust the decision policy $\pi$. In the simplest case, the algorithm may use trajectories of length $1$. Typically, an RL algorithm aims to optimize the policy to maximize expected discounted sum of rewards 
\begin{equation}\label{eq:discounted_rewards}
    G_t=\sum_{i=0}\gamma^i r_{t+i}
\end{equation} in each time step.

\subsection{Soft Actor-Critic}

Soft Actor-Critic \citep{haarnoja2017sac, haarnoja2017sacAA} is an online off-policy deep reinforcement learning algorithm based on the maximum entropy framework that aims to simultaneously maximize both the discounted sum of rewards and the entropy of the decision policy, thereby maximizing the expected soft-value function 
\begin{equation}
\begin{split}
    Q(a_t, s_t) &= r_t + \gamma\mathbb{E}_{s_{t+1}}[V(s_{t+1})]\\
    V(s_t) &=\mathbb{E}_{a\sim\pi}Q(s_t, a)+\alpha\widehat{\mathcal{H}}(\pi(\cdot|s_t))\\
    &= \mathbb{E}_{a\sim\pi}\left[Q(s_t, a) - \alpha\log\pi(a|s_t)\right]
\end{split}
\end{equation}
in each state, where $\widehat{\mathcal{H}}$ is the entropy estimator and $\alpha>0$ is the temperature parameter. It uses five neural network models: policy $\pi(a|s;\phi)$ parameterized by the weights vector $\phi$, and four critics that estimate the soft action-value function
$Q(a,s;\theta_1)$, $Q(a,s;\theta_2)$, $Q(a,s;\theta_{T,1})$, $Q(a,s;\theta_{T,2})$ parameterized by the weights vectors $\theta_1$, $\theta_2$, $\theta_{T,1}$, $\theta_{T,2}$. SAC uses two critic networks, $Q(a,s;\theta_1)$ and $Q(a,s;\theta_2)$ to mitigate positive bias in the policy improvement step \citep{hasselt2010doubleQL, fujimoto2018AddressingFA} and further two critic networks $Q(a,s;\theta_{T,1})$ and $Q(a,s;\theta_{T,2})$ as target networks with their weights being exponentially averaged weights $\theta_1$ and $\theta_2$ to further stabilize the training process \citep{mnih2015DQN}.

The temperature parameter $\alpha$ in the initial formulation of SAC was a hyperparameter \citep{haarnoja2017sac}. In later work, \citet{haarnoja2017sacAA} provided a method to compute the value of $\alpha$ to maximize the expected rewards by imposing a lower bound on the entropy $\mathcal{H}(\pi(\cdot|s))$.

\section{\uppercase{Related works}
}\label{sec:related_works}

\subsection{n-step returns}

To significantly reduce the variance of a Monte Carlo estimation of the discounted rewards' sum in equation \ref{eq:discounted_rewards} and to avoid the need to wait until the end of an episode to update the policy, a bootstrapping technique was introduced to estimate the value of the target return, namely 
\begin{equation}
    G^{(n)}_t=\sum_{i=0}^n\gamma^ir_t+\gamma^n v(s_{t+n})
\end{equation}
where $v(s_t)$ is a value-function approximation, 
\begin{equation}
    v(s) \cong \mathbb{E}_\pi\left(\left.\sum_{i=0}\gamma^ir_{t+i}\right|s_t=s\right)
\end{equation}
\citep{sutton1988TD}
However, this method introduces bias until the approximator $v(s_t)$ converges, resulting in a bias-variance trade-off based on the choice of $n$ \citep{kearns2000BiasVariance}. \citet{daley2024averagingNStep} further show that averaging n-step returns over multiple values of $n$ further decreases the variance of the return target. \citet{lyskawa2024acerac} suggest that n-step returns may improve the training results for large discount values.  

The n-step returns were successfully used in multiple deep RL methods, such as Actor-Critic with Experience Replay
\citep{wawrzynski2009ACER}, Asynchronous Advantage Actor-Critic \citep{mnih2016asynchronous}, and Rainbow \citep{hessel2018rainbow}.


\subsection{SAC with n-step returns}

There are prior works that approach the topic of using n-step returns with Soft Actor-Critic.

Improved Soft Actor-Critic \citep{shil2023improvedSAC} extends SAC by using n-step return estimation. Furthermore, to reduce the variance of entropy estimation, a singular policy network is replaced with dual delayed policy networks. However, the authors change the formulation of the maximum-entropy optimization by only considering the entropy of the policy in the last state in an n-step trajectory without providing any motivation for this change. Furthermore, the delayed policy network parameters are not used in Algorithm 1 listing in this paper, except for their update step, and the actor and critic parameters designations are used inconsistently across the paper. The SAC results on MuJoCo environments in this paper are significantly lower than those reported by the authors of the Soft Actor-Critic for both constant \citep{haarnoja2017sac} and variable entropy coefficient \citep{haarnoja2017sacAA}. Given the aforementioned errors in the paper and no publicly available source code, we consider the results reported for Improved Soft Actor-Critic to be irreproducible.

\subsection{Importance sampling in RL}

Training RL agents using data with a distribution different than the current one introduces bias. This problem is avoided in action-value function-based off-policy algorithms such as Deep Q-Network \citep{mnih2015DQN} and Soft Actor-Critic \citep{haarnoja2017sacAA} by using single-step return estimations and mitigated in Rainbow \citep{hessel2018rainbow} by using a small value of $n$ for n-step returns. \citet{wawrzynski2009ACER} solves the problem of bias introduced by using samples collected with a different policy using importance sampling \citep{kloek1978IS}, a weighting method that corrects the expected value estimation. This approach was later used by prioritized experience replay methods \citep{schaul2016per, horgan2018distributedPER, li2024per} to correct the bias introduced by the change of the probabilities of obtaining the samples in the experience buffer.

Importance sampling weights as used by \citet{wawrzynski2009ACER} can achieve arbitrarily large values and result in instability of the training process. To mitigate this problem, Actor-Critic with Experience Replay clips these weights to a value specified by a hyperparameter. This approach introduces a bias to the gradient estimator, which was addressed by \citet{wang2017seacer} by introducing a correction term based on the estimation of the expected returns.

Adaptive N-step Bootstrapping \citep{wang2021adaptiveNStepBootstrapping} proposes an approach for combining off-policy RL methods with n-step returns calculated using policies similar to the current one. It points out the bias introduced by using trajectories obtained using old policies. To mitigate this problem, the Adaptive N-step Bootstrapping limits the value of $n$ based on the trajectory age as an estimated measure of the difference between the old and the current policy.

High variance of importance sampling is also a challenge for the problem of stationary policy evaluation, which the authors of \citep{nachum2019DualDICE} attempt to correct using the DualDICE method. However, such methods of correction are not designed to correct estimates used in the n-step entropy framework, focusing on state-independent step rewards.

\section{\uppercase{SAC$n$: Soft Actor-Critic with n-step returns}}

Training of the $Q(s_t, a_t;\theta_i)$ networks using target networks $Q(s_t, a_t;\theta_{T,i})$ in SAC can be formulated as minimizing the loss function
\begin{equation}
    \mathcal{L}^{SAC}_t(\theta) = \sum_{i\in\{1,2\}}\left(Q(s_t, a_t; \theta_i) - R_t\right)^2
\end{equation}
where $\theta = [\theta_1^T, \theta_2^T]^T$
and $R_t$ is the soft action-value function target.

Base SAC uses a single time step to calculate the estimate of the soft action-value function target:
\begin{equation} \label{step1}
\begin{split}
    R_t = r_t + \gamma (&\alpha \widehat{\mathcal{H}}(\pi(\cdot|s_{t+1})) \\&+ \min_{i\in\{{1, 2\}}}Q(s_{t+1},\beta^\pi(s_{t+1});\theta_{T,i}))
\end{split}
\end{equation}
where $\beta^\pi(s_{t+1})$ is an action sampled from current policy $\pi$ and $\widehat{\mathcal{H}}(\pi(\cdot|s_{t+1}))$ is an estimator of the entropy $\mathcal{H}(\pi(\cdot|s_{t+1}))$. Specifically, SAC estimates the entropy of the action distribution as 
\begin{equation}
    \widehat{\mathcal{H}}(\pi(\cdot|s_{t+1})) = -\log\pi(\beta^\pi(s_{t+1})|s_{t+1})
\end{equation}

To introduce n-step returns, we may redefine the target $R_t$ to account for a trajectory of $n$ gathered experience samples: 
\begin{equation} \label{eq:stept}
\begin{split}
    {R^\tau_t} = &\sum_{i=0}^{\tau-1} \gamma^i r_{t+i} + \alpha\sum_{i=1}^\tau\gamma^i \widehat{\mathcal{H}}(\pi(\cdot|s_{t+i}))
    \\&+ \gamma^\tau \min_{i\in\{{1, 2\}}}Q(s_{t+\tau},\beta^\pi(s_{t+\tau});\theta_{T,i})\
\end{split}
\end{equation}

However, the above estimate is biased, because the actions $a_{t+1}, \dots, a_{t+\tau-1}$ that caused the environment to follow the sequence of states $s_t, \dots, s_{t+\tau}$ in off-policy algorithms such as SAC were taken according to the decision policy at that time, which may differ from the one that is being 
updated. To avoid bias introduced by using trajectories with actions sampled with any past policy, we apply importance sampling that allows for unbiased estimation of expected value using samples from another distribution:
\begin{equation} \label{eq:stept_is}
\begin{split}
    \mathcal{L}^\tau_t(\theta) = \sum_{i\in\{1,2\}}\left(Q(s_t, a_t; \theta_i) - {R^\tau}^*\right)^2\omega_\tau^\pi(t)
\end{split}
\end{equation}
where the importance sampling weight $\omega^\pi_\tau(t)$ is the ratio between the probability of sampling the actions $a_{t+1}, \ldots, a_{t+\tau-1}$ using the current policy $\pi$ and the probability with which these actions were actually sampled. Assuming that the actions at different time steps are drawn independently, such as in the original SAC, the importance sampling weight takes the form 
\begin{equation}\label{eq:is}
    \omega_\tau^\pi(t) =\begin{cases}\prod_{i=1}^{\tau-1}\frac{\pi(a_{t+i}|s_{t+i})}{\pi_{t+i}(a_{t+i}|s_{t+i})} & \mathrm{if}\;\tau > 1\\
    1 & \mathrm{if}\;\tau = 1
    \end{cases}
\end{equation}
where $\pi_{t}$ denotes the policy applied at the time step $t$.

For arbitrary policy distributions and $\tau > 1$, the importance sampling weight in \eqref{eq:is} is unbounded (see Subsection \ref{ssec:density_measurement} for analysis of the distribution of action probability density ratio). To ensure the stability of the algorithm, similarly to previous works that employ importance sampling to weight out-of-distribution experience samples, we employ clipping and scaling mechanisms, as used by \citet{wawrzynski2009ACER} and \citet{schaul2016per}, respectively.

Specifically, we calculate the weight of a sample $w_\tau(t)$ for each time step $t$ in batch $B$ as
\begin{equation}\label{eq:weight_scaling}
    w_\tau(t)=\frac{\min\{\omega^\pi_\tau(t), b\}}{\max_{t'\in B}\min\{\omega^\pi_\tau(t'), b\}}
\end{equation}
ensuring that both the weights are limited to $[0, 1]$ and that if near-infinite values of $\omega^\pi_\tau(t)$ occur in a batch, they do not reduce the other weights in this batch to 0.

To simplify the process of hyperparameter selection, we calculate the value of $b$ as a quantile of order $q_b$ of all the values of $\omega^\pi_\tau(t)$ for all $\tau \in \{1, \ldots, n\}$ in a batch. We selected the quantile approach because, typically, the number of samples with large values of $\omega^\pi_\tau(t)$ is relatively small, quantiles are immune to the scale of the outliers, and the quantile approach ensures that only a predetermined part of the trajectories in a batch will be clipped.

We take notice that SAC uses a stochastic estimation of the policy entropy $\mathcal{H}(\pi(\cdot|s))$. Using the same form of policy entropy estimation for a trajectory of $n$ steps would result in significantly increased variance contributed by the sum of discounted entropy estimates ${\sum_{i=0}^{\tau-1}}\gamma^i\alpha\widehat{\mathcal{H}}(\pi(\cdot|s_{t+i+1}))$: 
\begin{equation}
    \begin{split}
    Var&\left[{\sum_{i=0}^{\tau-1}}\gamma^i\alpha\widehat{\mathcal{H}}(\pi(\cdot|s_{t+i+1}))\right] \approx\\&\approx \sum_{i=0}^{\tau-1}\gamma^{2i}\alpha^2Var\left[\widehat{\mathcal{H}}(\pi(\cdot|s_{t+i+1}))\right].
    \end{split}
\end{equation}
Assuming that $Var\left[\widehat{\mathcal{H}}(\pi(\cdot|s')\right]\approx Var\left[\widehat{\mathcal{H}}(\pi(\cdot|s'')\right]$ for different states $s'$ and $s''$,
\begin{equation}
    \begin{split}
    Var&\left[{\sum_{i=0}^{\tau-1}}\gamma^i\alpha\widehat{\mathcal{H}}(\pi(\cdot|s_{t+i+1}))\right] \approx \\
    &\approx
    \begin{cases}
        \tau\alpha^2Var\left[\widehat{\mathcal{H}}(\pi(\cdot|s_{t+1}))\right] \;\mathrm{if}\; \gamma=1 \\
        \frac{1-\gamma^{2\tau}}{1-\gamma^2}\alpha^2Var\left[\widehat{\mathcal{H}}(\pi(\cdot|s_{t+1}))\right]\;\mathrm{otherwise,}
    \end{cases}
    \end{split}
\end{equation}
so the variance contributed by the policy entropy estimates is increased approximately $k(\tau)$ times when compared to the base Soft Actor-Critic, where
\begin{equation}
    k(\tau)=\begin{cases}
        \tau\;\mathrm{if}\; \gamma=1 \\
        \frac{1-\gamma^{2\tau}}{1-\gamma^2}\;\mathrm{otherwise}
    \end{cases}
\end{equation}
To reduce this variance, we propose $\tau$-sampled entropy estimation
\begin{equation}
    \widehat{\mathcal{H}}^{(\tau)}(\cdot|\pi(s)) = \frac{1}{\lfloor k(\tau)\rceil}\sum_{j=1}^{\lfloor k(\tau)\rceil}-\log\pi (\beta_j^\pi(s)|s)
\end{equation}
which uses $\lfloor k(\tau)\rceil$ (so $k(\tau)$ rounded to the nearest integer) different independently sampled actions $\beta^\pi_j(s_{t+1})$, indexed by $j$,
instead of just a single one, reducing its variance when compared to just using $\widehat{\mathcal{H}}(\pi(\cdot|s))$:
\begin{equation}
    Var\left[\widehat{\mathcal{H}}^{(\tau)}(\pi(\cdot|s))\right] = \frac{1}{\lfloor k(\tau)\rceil}Var\left[\widehat{\mathcal{H}}(\pi(\cdot|s))\right]
\end{equation}
and reducing the variance contributed by the entropy estimation (under the previous assumptions) to approximately the same value as the variance of the entropy estimation in SAC
\begin{equation}
    \begin{split}
    Var&\left[{\sum_{i=0}^{\tau-1}}\gamma^i\alpha\widehat{\mathcal{H}}^{(\tau)}(\pi(\cdot|s_{t+i+1}))\right] \approx\\&\approx 
    \frac{k(\tau)}{\lfloor k(\tau)\rceil}
    \alpha^2Var\left[\widehat{\mathcal{H}}(\pi(\cdot|s_{t+1}))\right]
    \end{split}
\end{equation}

All of the aforementioned changes result in the following loss $\mathcal{L}^{(n)}_t(\theta)$ for the SAC$n$
\begin{equation}
    \begin{split}
        \mathcal{L}^{(n)}_t(\theta) &= \frac{1}{n}\sum_{\tau=1}^n\mathcal{L}_t^\tau(\theta) \\
        \mathcal{L}^{\tau}_t(\theta) &= \sum_{i\in\{1,2\}}\left(Q(s_t, a_t; \theta_i) - R^\tau_t\right)^2w_\tau(t) \\
        R^\tau_t&=\sum_{i=0}^{\tau-1}\gamma^i\left(r_{t+i}+\gamma\widehat{\mathcal{H}}^{(\tau)}(\pi(\cdot|s_{t+i+1}))\right)\\
        &+ \gamma^\tau \min_{i\in\{{1, 2\}}}Q(s_{t+\tau},\beta^\pi(s_{t+\tau});\theta_{T,i})
    \end{split}
\end{equation}
as a mean of all partial losses weighted by $w_\tau(t)$ over $\tau \in\{1,\ldots,n\}$ to ensure sample relevance even for low values of $w_\tau(t)$ by leveraging that typically lower values of $\tau$ should result in $w_\tau(t)$ closer to 1.

Note that the replay buffer has to include the probability density $\pi_t(a_t|s_t)$ of each action at the time step of its selection $t$ for the calculation of probability density ratio $\frac{\pi(a_t|s_t)}{\pi_t(a_t|s_t)}$.

\section{\uppercase{Experimental study}}\label{sec:experimental_study}

\subsection{Setup}

We performed our experiments on five environments using MuJoCo simulator \citep{todorov2012mujoco}: Ant, HalfCheetah, Hopper, Swimmer, and Walker2d.

We used the optimized hyperparameters for SAC available in the RL Baselines3 Zoo repository \citep{raffin2020rlzoo3}, which are the same as those used in \citet{haarnoja2017sacAA} except for the number of timesteps. For SAC$n$, we kept parameters used by SAC where possible and tested five different values of $n$: 2, 4, 8, 16, and 32. We set $q_b=0.75$. Full list of hyperparameters is available in appendix C.

Each run lasted $10^6$ time steps. Every $10^4$ time steps, the training was paused, and we measured the performance of the agent over 5 test runs. Each experiment was repeated 9 times unless noted otherwise.

We measured the values of action probability ratio $\frac{\pi(a_t|s_t)}{\pi_t(a_t|s_t)}$ for the vanilla SAC algorithm to analyze the expected behavior of SAC$n$ and to underline the need for a well-designed stabilizing mechanism.

We also provide an experimental evaluation of the influence of different hyperparameters and components of the SAC$n$ algorithm on its performance. We ran the experiments with $q_b$ decreased to $0.5$ and $0.25$ and increased to $0.85$ and $0.95$ for each $n$ in $\{2, 4, 8\}$. We also present results for SAC$n$ without $\tau$-sampled entropy estimation and for SAC with entropy estimation changed to the $\tau$-sampled entropy estimation for three different values of $\tau$: 2, 4, 8. 

The source code used for experiments reported in this section is available online as a Git repository\footnote{\url{https://gtihub.com/lychanl/sacn}}.

\subsection{Probability density ratio measurement}\label{ssec:density_measurement}

\begin{figure}
    \centering
    \includegraphics[width=0.75\linewidth]{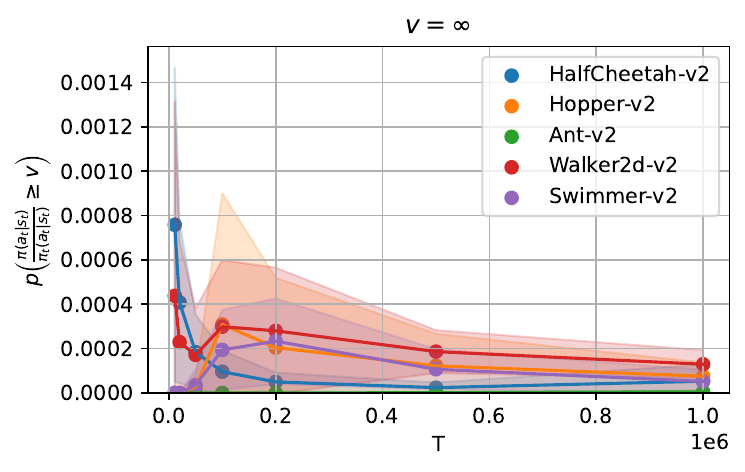}
    \caption{Part of the samples that result in infinite action probability densities due to the numerical precision of floating point numbers typically used in machine learning.}
    \label{fig:density_inf_intext}
\end{figure}

We measured the probability density ratios for each batch used for training the agent during the $1000$-step periods starting at the time steps 10001, 19001, 49001, 199001, 499001, and 999001 to capture the distributions of the probability ratios during different parts of the training. The aggregated values were averaged over 5 runs.

Figure \ref{fig:density_inf_intext} shows the part of the measured density ratios that, due to the precision of 32-bit floating point numbers, are infinite. These values, if used for calculating the importance sampling weights without any clipping or other stabilizing mechanism, would cause the instability of the training process. This shows the need for such safeguards in the algorithm itself.

Even if a large value of probability a weight does not exceed the numerical precision of data format, it could destabilize the training process if used without any clipping or scaling, or, with scaling, the weights of other samples would be so small that these samples would not contribute to the training process.

The occurrence of such values underlines the need for a stabilizing mechanism.

Results for different clipping values are presented in appendix A.

\subsection{Results}

\begin{figure*}
    \centering
    \includegraphics*[width=0.32\linewidth]{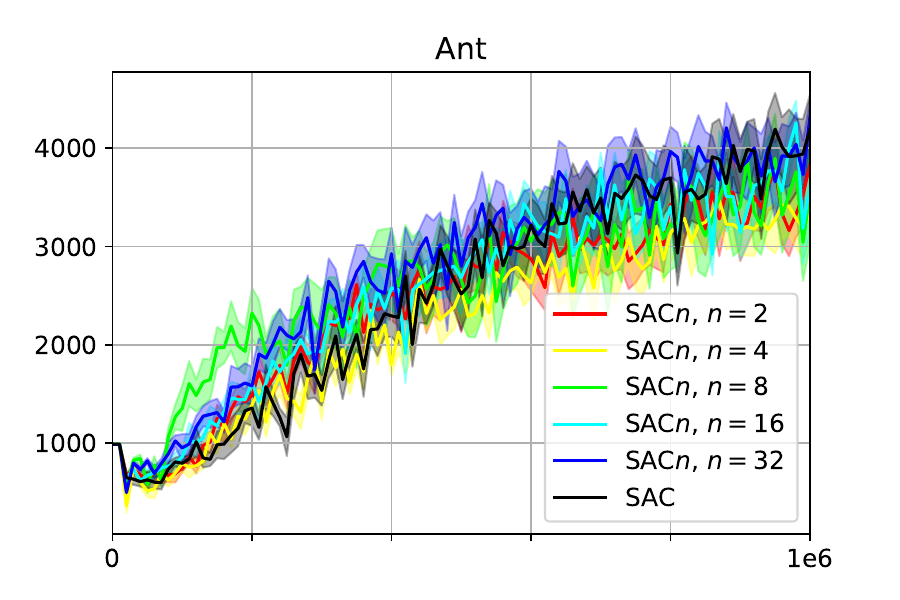}
    \includegraphics*[width=0.32\linewidth]{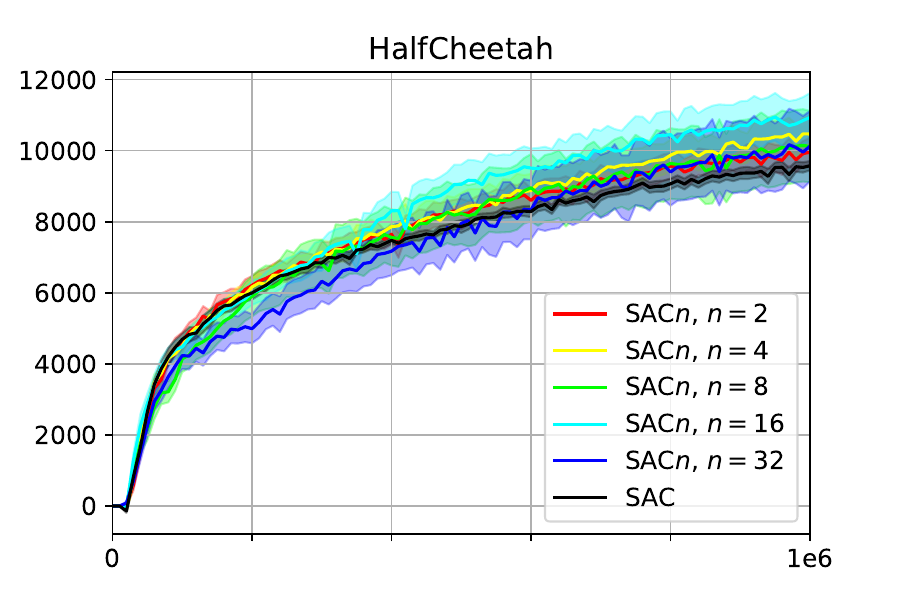}
    \includegraphics*[width=0.32\linewidth]{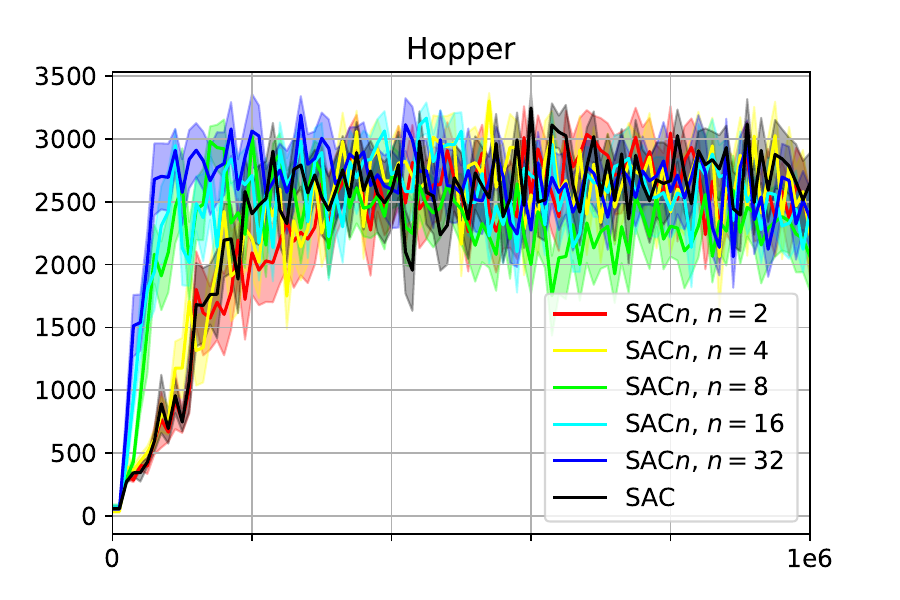}\\
    \includegraphics*[width=0.32\linewidth]{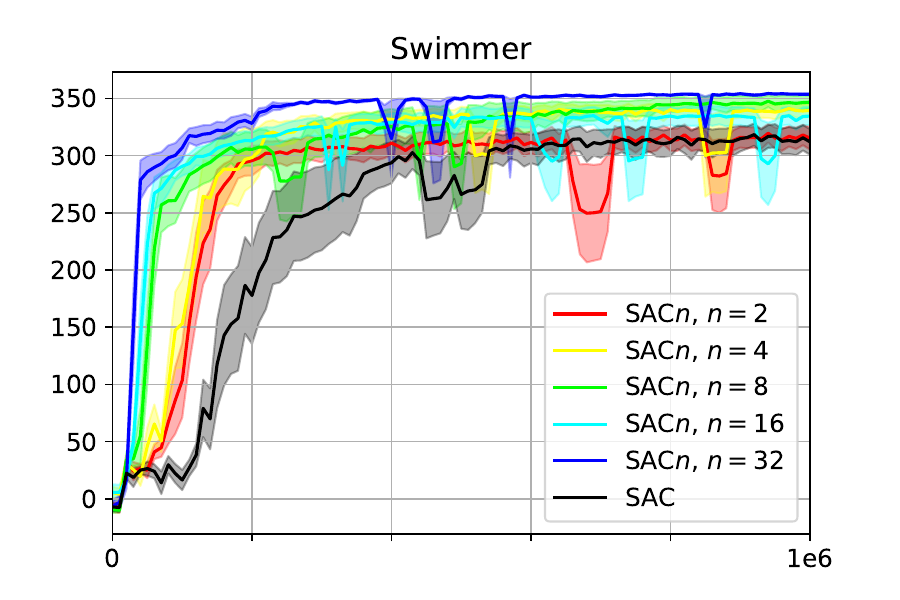}
    \includegraphics*[width=0.32\linewidth]{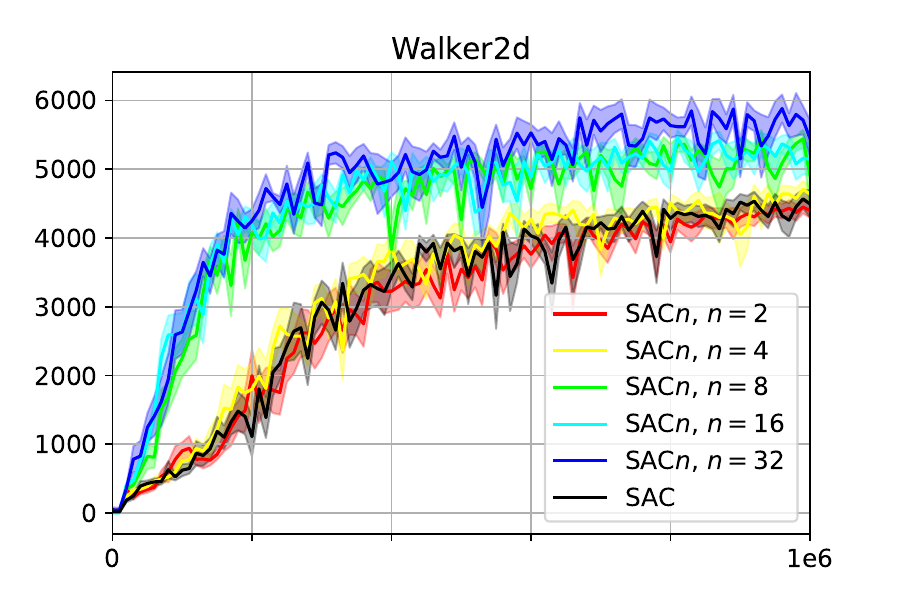}
    \caption{Results for MuJoCo environments for SAC$n$ with $n\in\{2, 4, 8\}$ and SAC.}
    \label{fig:results}
\end{figure*}

\begin{table}
    \centering
    \caption{Evaluation results averaged over last $3\cdot10^4$ time steps, their standard errors and the probability $p$ that the results for SAC$n$ are greater than or equal to the results for SAC according to the Welch's t-test. Best scores for each environment and values of $p>0.9$ are in boldface.}\label{tab:measurements}

    \begin{tabular}{l|l|r|r}
        \hline
        Environment & $n$ & SAC$n$ & $p$ \\
        \hline
Ant & 1 (SAC) & $\mathbf{3941\pm318}$ & -- \\
& 2 & $3450\pm233$ & 0.05 \\
& 4 & $3265\pm214$ & 0.01 \\
& 8 & $3509\pm343$ & 0.12 \\
& 16 & $3632\pm189$ & 0.10 \\
& 32 & $\mathbf{3908\pm265}$ & 0.36 \\
\hline
HalfCheetah & 1 (SAC) & $9439\pm147$ & -- \\
& 2 & $9804\pm363$ & 0.89 \\
& 4 & $\mathbf{10325\pm453}$ & \textbf{0.97} \\
& 8 & $10016\pm1008$ & 0.75 \\
& 16 & $\mathbf{10808\pm671}$ & \textbf{0.97} \\
& 32 & $9939\pm992$ & 0.73 \\
\hline
Hopper & 1 (SAC) & $\mathbf{2711\pm140}$ & -- \\
& 2 & $2536\pm197$ & 0.30 \\
& 4 & $\mathbf{2629\pm213}$ & 0.44 \\
& 8 & $2287\pm222$ & 0.09 \\
& 16 & $2447\pm206$ & 0.20 \\
& 32 & $2517\pm169$ & 0.25 \\
\hline
Swimmer & 1 (SAC) & $315\pm11$ & -- \\
& 2 & $316\pm9$ & 0.27 \\
& 4 & $339\pm6$ & \textbf{0.92} \\
& 8 & $\mathbf{346\pm6}$ & \textbf{0.97} \\
& 16 & $324\pm11$ & 0.49 \\
& 32 & $\mathbf{354\pm0}$ & \textbf{1.00} \\
\hline
Walker2d & 1 (SAC) & $4442\pm144$ & -- \\
& 2 & $4376\pm111$ & 0.37 \\
& 4 & $4503\pm170$ & 0.60 \\
& 8 & $5202\pm128$ & \textbf{1.00} \\
& 16 & $5229\pm138$ & \textbf{1.00} \\
& 32 & $\mathbf{5606\pm171}$ & \textbf{1.00} \\
\hline
    \end{tabular}
\end{table}

Figure~\ref{fig:results} presents learning curves for SAC and SAC$n$, averaged over all runs, which are presented together with their standard errors. Table~\ref{tab:measurements} contains the results, averaged over evaluation runs during the last $3\cdot10^4$ time steps. This table also includes probabilities, calculated using Welch's t-test, that results obtained by SAC$n$ on a given environment for each $n$ are greater than or equal to the results obtained by the base SAC algorithm.

Our method shows significant improvement over the base SAC algorithm for three out of five used environments: HalfCheetah, Swimmer, and Walker2d. For Ant and Hopper, it obtains similar results as SAC.

For most environments, as seen in the plotted learning curves, larger values of $n$ allow SAC$n$ to obtain better results. It is likely caused by longer trajectories being less dependent on the critic network. As the $n$-step approach reduces the bias introduced by the critic estimating the returns of past policies. This is especially evident for the Swimmer environment, which has the highest discount factor of $\gamma=0.999$ as compared to $\gamma=0.99$ used for other environments, where even $n=2$ allows for quicker agent training, and larger $n$ results in faster training. However, larger values of $n$ increase the amount of calculations required for each update step - the number of neural network forward passes is in $O(n)$, as each time step in a trajectory requires forward passes of the target critic model and the actor model. The $\tau$-sampled entropy estimation requires sampling of $n$ actions for each time step of a trajectory of length $n$ (we reuse sampled actions $\beta^\pi_j(\pi(\cdot|s_t), j\in\{1, \ldots, n\}$ for each entropy estimator $\widehat{\mathcal{H}}^{(\tau)}(\pi(\cdot|s))$ where $j \leq\tau$), pushing the algorithm into $O(n^2)$. However, the policy parameters may be reused between different sampled actions and for relatively small values of $n$, such as the ones used in this study, the amount of calculations required by $\widehat{\mathcal{H}}^{(\tau)}(\pi(\cdot|s))$ is much lower than the number of calculations required by a singular forward pass of a neural network.

\subsection{Ablation study}\label{ssec:ablation}
\begin{table*}
    \centering
    \caption{Results for SAC$n$ for different values of $q_b$ and $n$. Results for base SAC are included for reference. Best values for each environment-$n$ pair are in boldface.}
    \label{tab:c_ablation}
    \begin{tabular}{l|l|lll|l}
\hline
Environment & $q_b$ & $n=2$ & $n=4$ & $n=8$ & SAC \\
\hline
Ant & $0.25$ & $\mathbf{4343\pm298}$ & $3732\pm278$ & $3832\pm274$ & $3941\pm318$ \\
& $0.5$ & $3392\pm275$ & $\mathbf{4990\pm178}$ & $3243\pm158$ & \\
& $0.75$ & $3450\pm233$ & $3265\pm214$ & $3509\pm343$ & \\
& $0.85$ & $3638\pm251$ & $4033\pm239$ & $\mathbf{4410\pm323}$ & \\
& $0.95$ & $3661\pm288$ & $3523\pm232$ & $3483\pm355$ & \\
\hline
HalfCheetah & $0.25$ & $8989\pm400$ & $\mathbf{10078\pm963}$ & $\mathbf{10519\pm290}$ & $9439\pm147$ \\
& $0.5$ & $8490\pm466$ & $8419\pm1179$ & $7895\pm1401$ & \\
& $0.75$ & $9804\pm363$ & $\mathbf{10325\pm453}$ & $\mathbf{10016\pm1008}$ & \\
& $0.85$ & $\mathbf{9718\pm468}$ & $9550\pm546$ & $9701\pm1121$ & \\
& $0.95$ & $9580\pm417$ & $9133\pm562$ & $7558\pm878$ & \\
\hline
Hopper & $0.25$ & $\mathbf{3193\pm122}$ & $\mathbf{2799\pm150}$ & $2660\pm194$ & $2711\pm140$ \\
& $0.5$ & $\mathbf{3196\pm139}$ & $2457\pm170$ & $2240\pm141$ & \\
& $0.75$ & $2536\pm197$ & $2629\pm213$ & $2287\pm222$ & \\
& $0.85$ & $2396\pm270$ & $2626\pm220$ & $\mathbf{2859\pm105}$ & \\
& $0.95$ & $2528\pm117$ & $\mathbf{2774\pm171}$ & $\mathbf{2759\pm201}$ & \\
\hline
Swimmer & $0.25$ & $322\pm7$ & $322\pm8$ & $333\pm7$ & $315\pm11$ \\
& $0.5$ & $296\pm24$ & $329\pm7$ & $324\pm10$ & \\
& $0.75$ & $316\pm9$ & $\mathbf{339\pm6}$ & $\mathbf{346\pm6}$ & \\
& $0.85$ & $322\pm8$ & $324\pm8$ & $\mathbf{342\pm5}$ & \\
& $0.95$ & $\mathbf{332\pm7}$ & $\mathbf{343\pm5}$ & $336\pm8$ & \\
\hline
Walker2d & $0.25$ & $4544\pm176$ & $\mathbf{5399\pm125}$ & $4539\pm167$ & $4442\pm144$ \\
& $0.5$ & $4412\pm160$ & $4614\pm122$ & $5098\pm122$ & \\
& $0.75$ & $4376\pm111$ & $4503\pm170$ & $\mathbf{5202\pm128}$ & \\
& $0.85$ & $4240\pm202$ & $4527\pm133$ & $4697\pm148$ & \\
& $0.95$ & $\mathbf{4833\pm145}$ & $4545\pm136$ & $4480\pm187$ & \\
\hline
    \end{tabular}
\end{table*}

\begin{table*}
    \centering
    \caption{Results for SAC$n$ without $\tau$-sampled entropy estimation. Results for SAC$n$ with $\tau$-sampled entropy estimation and for base SAC are included for reference.}
    \label{tab:sacn_no_tau_entropy}
    \begin{tabular}{l|l|lll|l}
\hline
Environment & variant & $n=2$ & $n=4$ & $n=8$ & SAC \\
\hline
Ant & SACn without $\widehat{\mathcal{H}}^{(\tau)}$
& $2935\pm109$ & $3559\pm272$ & $3947\pm340$ & $3941\pm318$ \\
& SACn with $\widehat{\mathcal{H}}^{(\tau)}$
& $3450\pm233$ & $3265\pm214$ & $3509\pm343$ & \\
\hline
HalfCheetah & SACn without $\widehat{\mathcal{H}}^{(\tau)}$
& $9279\pm345$ & $7616\pm972$ & $7733\pm1274$ & $9439\pm147$ \\
& SACn with $\widehat{\mathcal{H}}^{(\tau)}$
& $9804\pm363$ & $10325\pm453$ & $10016\pm1008$ & \\
\hline
Hopper & SACn without $\widehat{\mathcal{H}}^{(\tau)}$
& $2954\pm221$ & $2752\pm359$ & $2618\pm241$ & $2711\pm140$ \\
& SACn with $\widehat{\mathcal{H}}^{(\tau)}$
& $2536\pm197$ & $2629\pm213$ & $2287\pm222$ & $2711\pm140$ \\
\hline
Swimmer & SACn without $\widehat{\mathcal{H}}^{(\tau)}$
& $304\pm10$ & $332\pm9$ & $342\pm7$ & $315\pm11$ \\
& SACn with $\widehat{\mathcal{H}}^{(\tau)}$
& $316\pm9$ & $339\pm6$ & $346\pm6$ & $315\pm11$ \\
\hline
Walker2d & SACn without $\widehat{\mathcal{H}}^{(\tau)}$
& $4213\pm185$ & $4797\pm96$ & $5005\pm152$ & $4442\pm144$ \\
& SACn with $\widehat{\mathcal{H}}^{(\tau)}$
& $4376\pm111$ & $4503\pm170$ & $5202\pm128$ & $4442\pm144$ \\
\hline
    \end{tabular}
\end{table*}

\begin{table*}
    \centering
    \caption{Results for SAC with $\tau$-sampled entropy estimation. Results for base SAC are included for reference.}
    \label{tab:sac_tau_entropy}
    \begin{tabular}{l|lll|l}
\hline
Environment & $\tau=2$ & $\tau=4$ & $\tau=8$ & SAC \\
\hline
Ant & $3872\pm306$ & $4239\pm282$ & $3786\pm272$ & $3941\pm318$ \\
\hline
HalfCheetah & $8911\pm250$ & $9393\pm267$ & $8592\pm387$ & $9439\pm147$ \\
\hline
Hopper & $2732\pm180$ & $2881\pm178$ & $2905\pm112$ & $2711\pm140$ \\
\hline
Swimmer & $325\pm9$ & $329\pm7$ & $338\pm9$ & $315\pm11$ \\
\hline
Walker2d & $4205\pm89$ & $4436\pm157$ & $4465\pm90$ & $4442\pm144$ \\
\hline
    \end{tabular}
\end{table*}

We performed an ablation study to investigate the influence of the hyperparameter $q_b$ on the results obtained by SAC$n$ (results in Table \ref{tab:c_ablation}) and the effects of using $\tau$-sampled entropy estimation $\widehat{\mathcal{H}}^{(\tau)}(\pi(\cdot|s))$ by running SAC$n$ without the $\tau$-sampled entropy estimation (results in Table \ref{tab:sacn_no_tau_entropy}) and by running SAC with $\tau$-sampled entropy estimation (results in \ref{tab:sac_tau_entropy}).

We found that SAC$n$ for most environments and values of $n$ works well for the intermediate value of $q_b=0.75$. Results for lower values of $q_b\in\{0.25, 0.5\}$ are closer to SAC, although they offer improvement for some environment-$n$ pairs. Similarly, setting $q_b$ to the high values of $0.85$ and $0.95$ may improve the obtained results, depending on the environment. In general, we find that whether SAC$n$ achieves better results than SAC or not is, outside of a few cases, consistent. However, setting $q_b$ to even lower values would result in the algorithm clipping all weights to a single value and failing to reduce $n$-step estimation bias. On the other hand, setting $q_b$ to a value of 1 would result in loss estimations for samples with very large values of probability density ratio dominating the averaged loss estimation for a given value of $\tau$.

Our experiments confirmed that the influence of the variance of the entropy estimation is significant when using $n$-step returns. Changing the entropy estimation method for the single-step SAC to $\tau$-sampled entropy estimation $\widehat{\mathcal{H}}^{(\tau)}$ did not result in large changes to the training process. However, SAC$n$ obtains similar or lower results when using single-sample estimations, especially for the HalfCheetah environments, suggesting that removing the $\tau$-sampled entropy estimation has a negative effect on the stability of SAC$n$.

Overall, our ablation study confirms that a more stable entropy estimation allows for improvement of the results for $n$-step returns in the maximum entropy framework and shows that the improvement of the results of SAC$n$ over SAC is not caused only by the entropy estimation method. The ablation study also suggests that the $q_b$ hyperparameter should be set to intermediate values, and although best values of $q_b$ vary between the environment and the value of $n$, in most cases SAC$n$ either outperforms SAC or achieves similar scores.

Full results are presented in appendix B.

\section{\uppercase{Summary}}

In this paper, we have introduced SAC$n$, Soft Actor-Critic with n-step returns, an algorithm that combines the maximum entropy framework used by Soft Actor-Critic with n-step return estimation. We provided an analysis of the variance of the entropy estimator that shows the increase in the variance of the entropy estimation for n-step targets. To reduce this variance, we proposed the $\tau$-sampled entropy estimator.

We verified experimentally the quality of the proposed solution using MuJoCo simulated robotic environments. Using Welch's t-test, we showed that SAC$n$ offers significant improvements for most of the used environments. Furthermore, even if SAC$n$ does not improve the final result, it often allows an increase in the speed of training in the beginning.

We experimentally confirmed the negative effect of the increased variance of entropy estimation when using $n$-step returns, and we provided a modified entropy estimation method that mitigates this problem.

Overall, we recommend using SAC$n$ instead of SAC for the environments with time-consuming environment steps, where the additional computations required by SAC$n$ have the lowest impact on the overall time needed to train the agent, and in settings that require very large values of $\gamma$, to reduce the dependency on the target critic network and the bias introduced by the delay in its training. When using SAC$n$, we recommend setting its hyperparameter $q_b$ to an intermediate value, and we showed in this work that $q_b=0.75$ performs well in the tested environments.


\section*{\uppercase{Acknowledgements}}

We gratefully acknowledge Polish high-performance computing infrastructure PLGrid (HPC Center: CI TASK) for providing computer facilities and support within computational grant nos. PLG/2024/017232 and PLG/2024/017692.


\bibliographystyle{apalike}
\small{\bibliography{bibliography}}

\appendix
\section{\uppercase{Density measurement}}\label{app:density_mearuements}

\begin{figure*}
    \centering
    \includegraphics[width=0.24\linewidth]{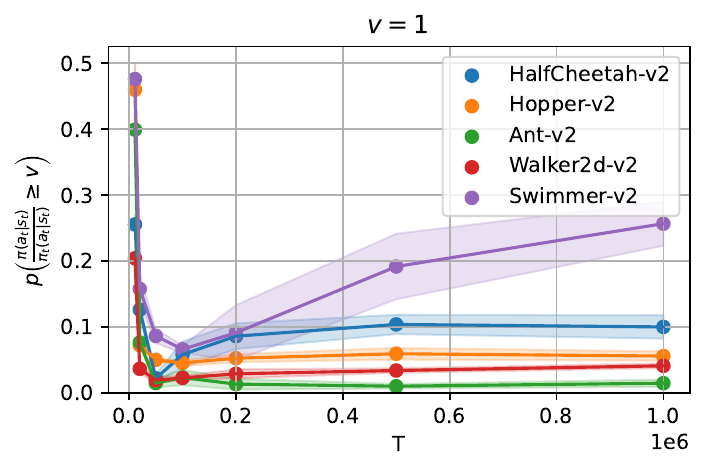}
    \includegraphics[width=0.24\linewidth]{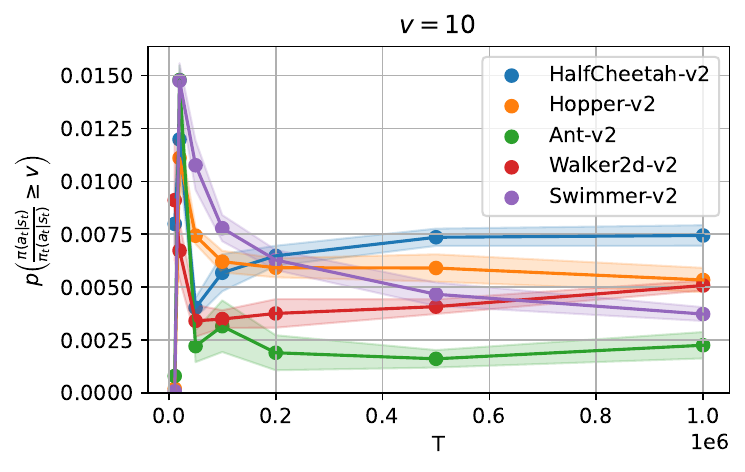}
    \includegraphics[width=0.24\linewidth]{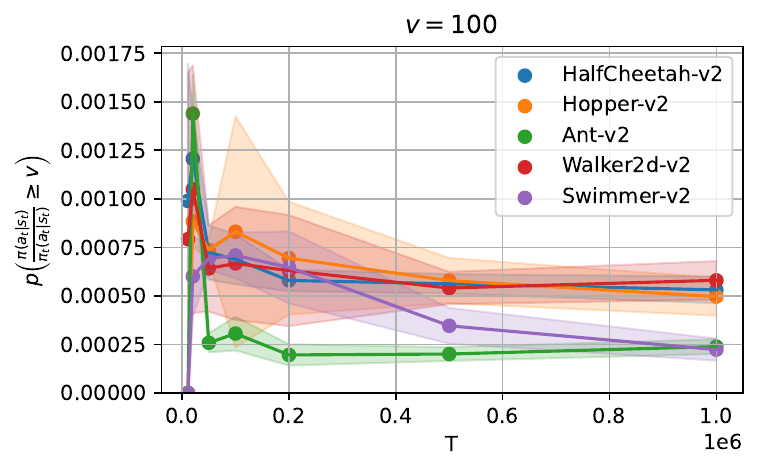}
    \includegraphics[width=0.24\linewidth]{plots/plot_sac_density_vinf.pdf}
    \caption{Part of samples that result in action probability densities equal to or exceeding given threshold value. Infinity values are the result of numerical precision of 32-bit floating point numbers typically used in machine learning.}
    \label{fig:app_density_measurements}
\end{figure*}

Figures \ref{fig:app_density_measurements} present which part of the probability density ratios are equal to or greater than given threshold values, measured for batches used for training the base SAC agent during the $1000$-step periods at the time steps \{10001, \dots, 11000\}, \{19001, \dots, 20000\}, \{49001, \dots, 50000\}, \{99001, \dots, 100000\}, \{199001, \dots, 200000\}, \{499001, \dots, 500000\}, \{999001, \dots, 1000000\}, to capture the distributions of the probability ratios during different parts of the training. The aggregated values were averaged over 5 runs.

The measured action probability density ratios allow to formulate the following observations:
\begin{enumerate}
    \item Significantly below half of the samples in the buffer have a probability density ratio of 1 or greater, except during the very first steps of the training. It means that for the most of the samples in the buffer, the probability of the selected action decreased. This is likely caused by most of the explored actions resulting in suboptimal returns and the policy being fitted to match only the small of best actions.
    \item During the whole training process samples with large importance sampling values keep occurring, and some of them, due to the precision of number representation, result in infinite probability density values. As the policy is likely fitted to replicate a small part of the actions selected by the exploration process, this underlines the need to properly handle such samples.
\end{enumerate}

\section{\uppercase{Ablation results}}\label{app:ablation}

\begin{figure*}
    \centering
    \includegraphics[width=0.21\linewidth]{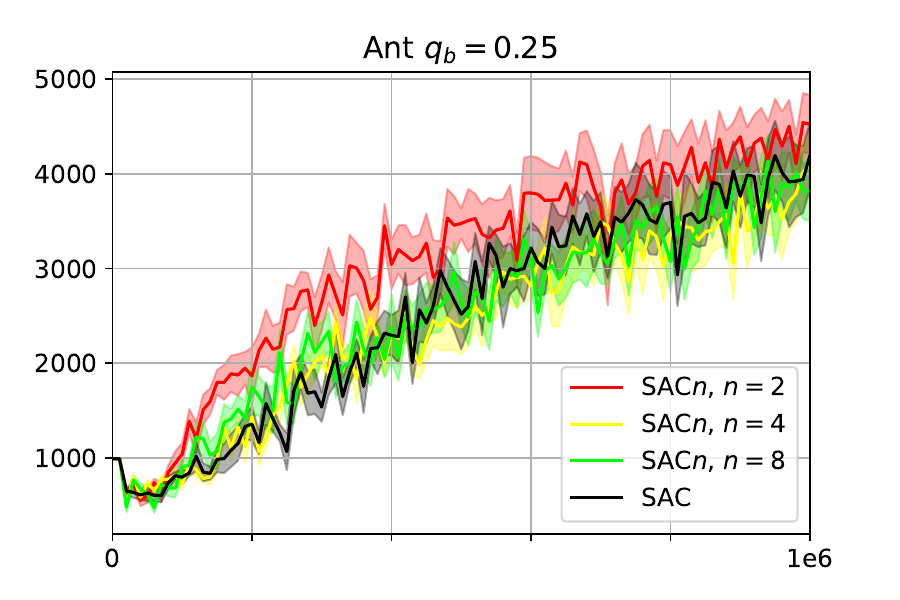}
    \hspace{-0.5cm}
    \includegraphics[width=0.21\linewidth]{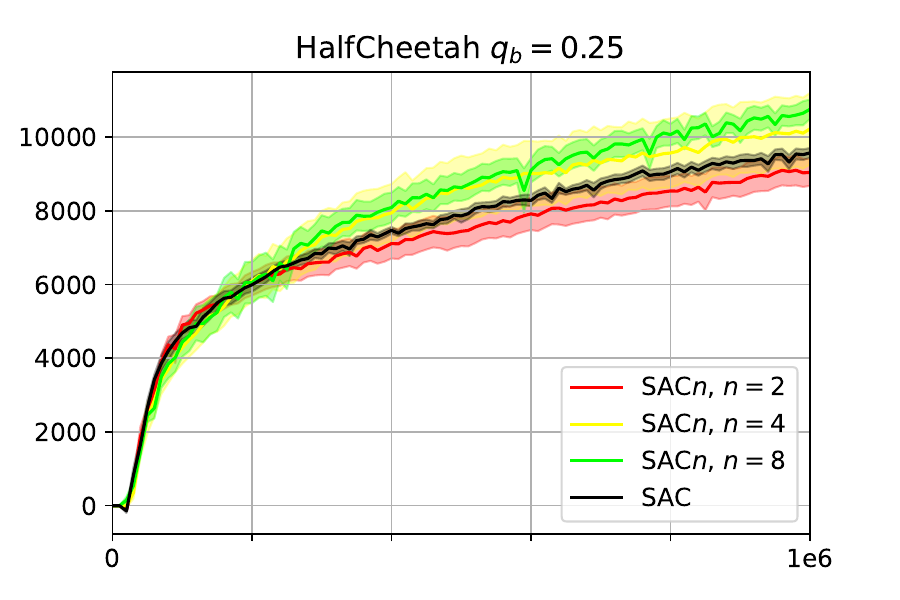}
    \hspace{-0.5cm}
    \includegraphics[width=0.21\linewidth]{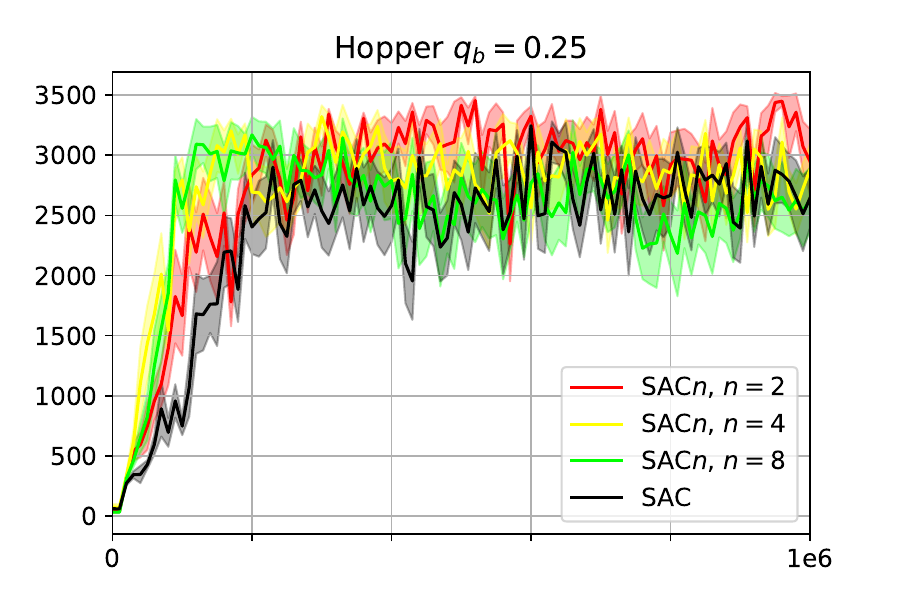}
    \hspace{-0.5cm}
    \includegraphics[width=0.21\linewidth]{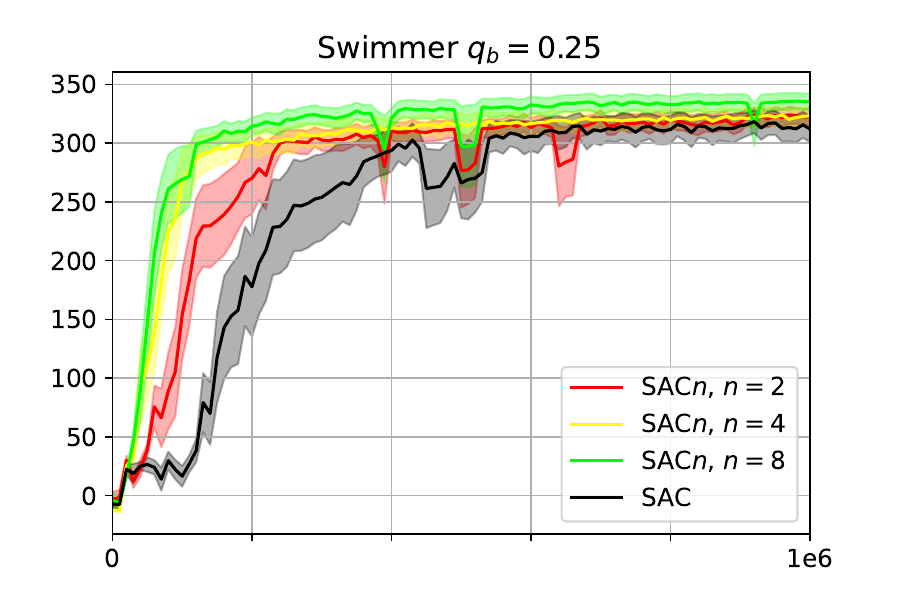}
    \hspace{-0.5cm}
    \includegraphics[width=0.21\linewidth]{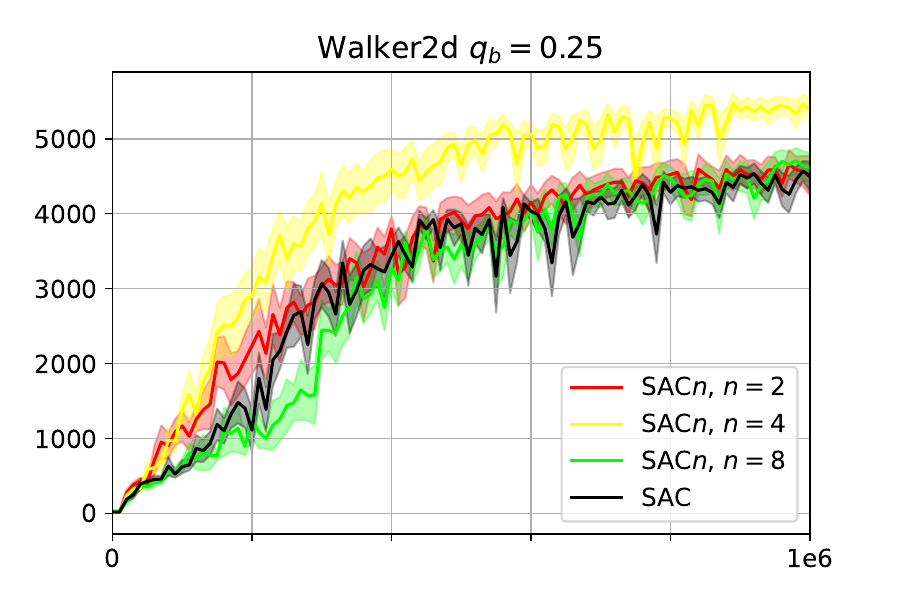}
    \hspace{-0.5cm}\\
    \includegraphics[width=0.21\linewidth]{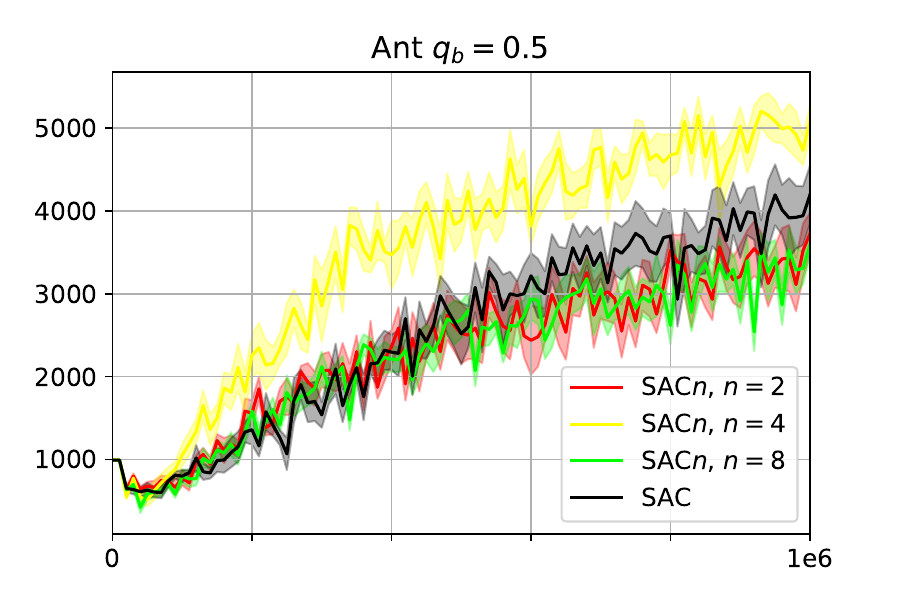}
    \hspace{-0.5cm}
    \includegraphics[width=0.21\linewidth]{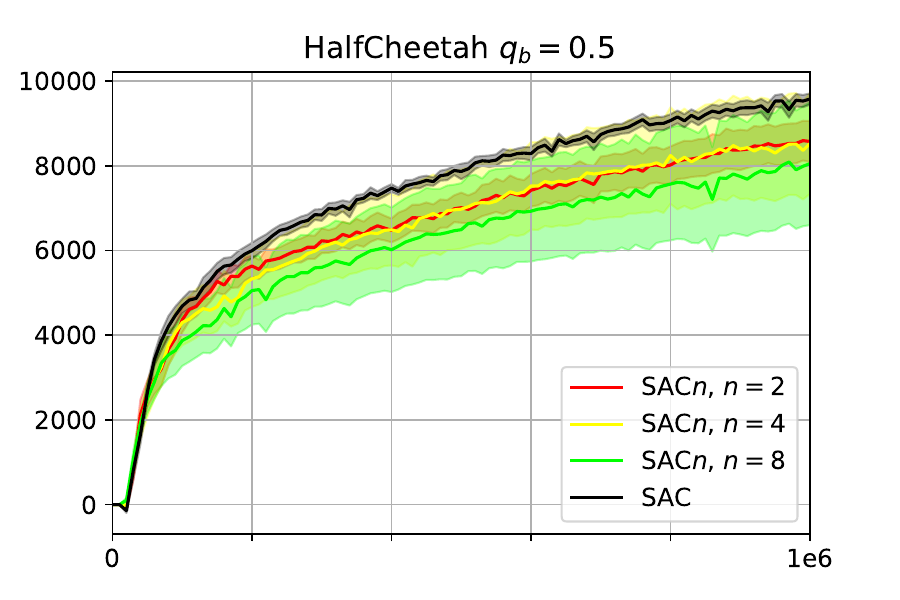}
    \hspace{-0.5cm}
    \includegraphics[width=0.21\linewidth]{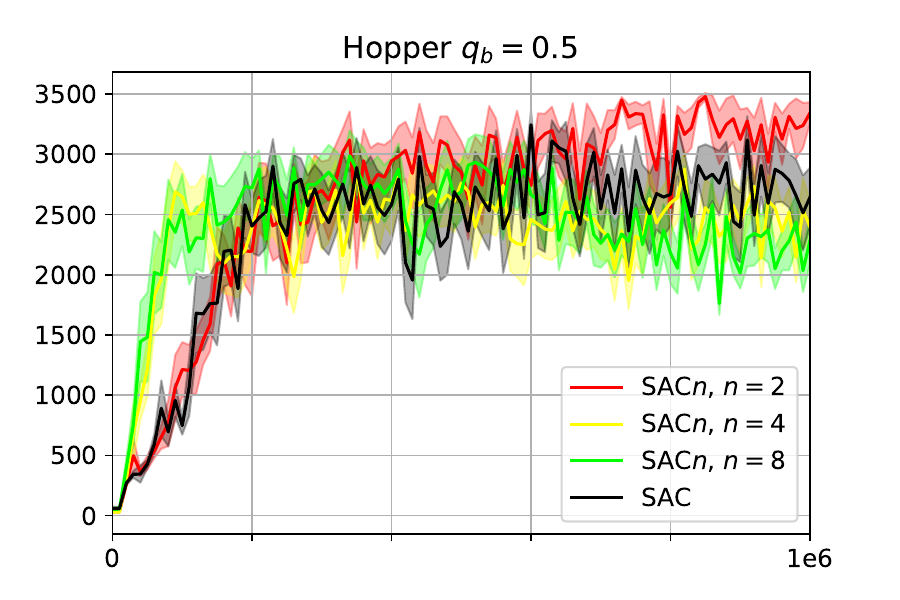}
    \hspace{-0.5cm}
    \includegraphics[width=0.21\linewidth]{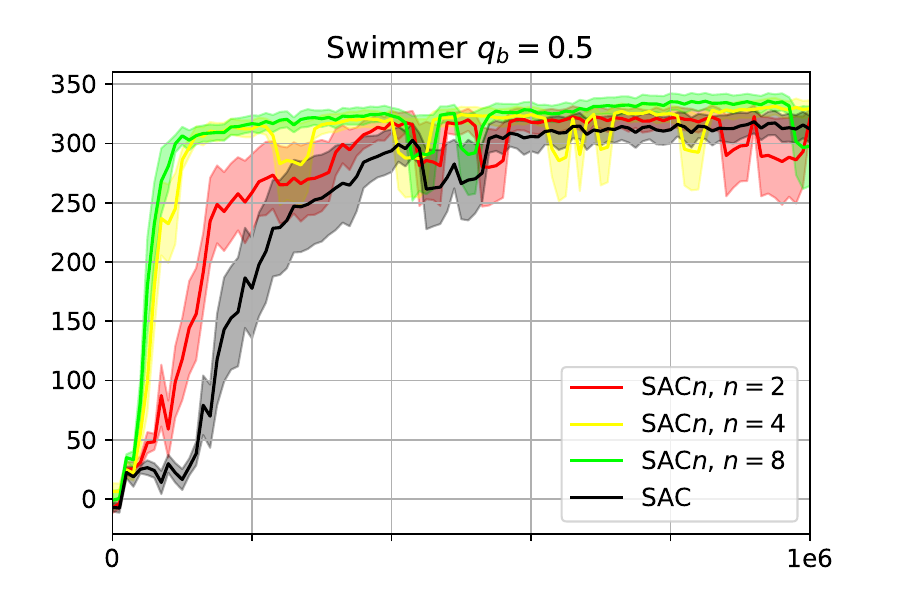}
    \hspace{-0.5cm}
    \includegraphics[width=0.21\linewidth]{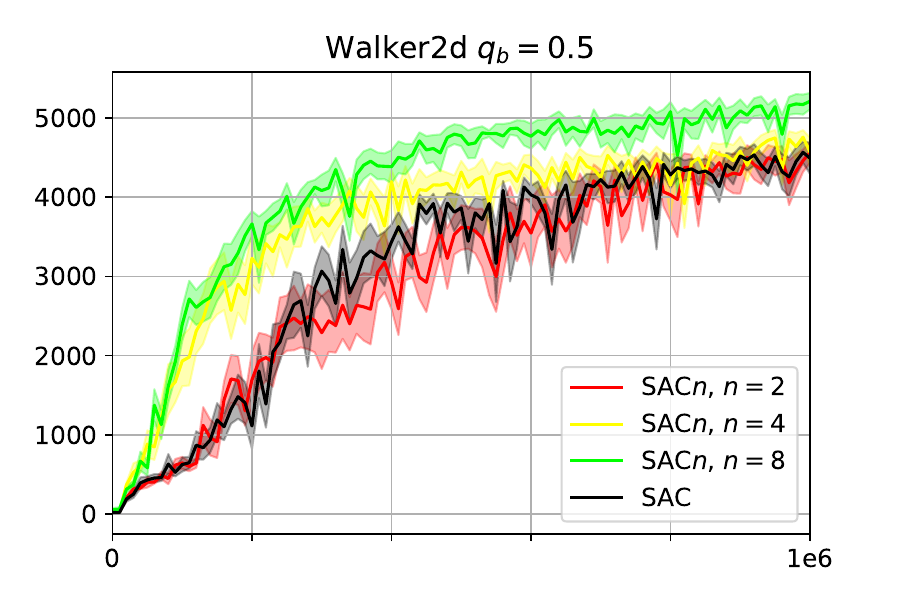}
    \hspace{-0.5cm}\\
    \includegraphics[width=0.21\linewidth]{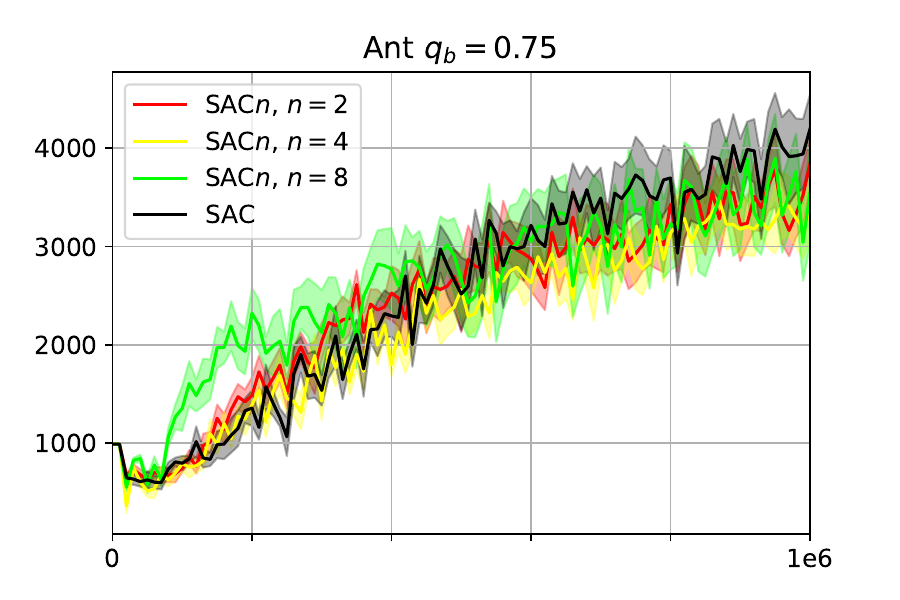}
    \hspace{-0.5cm}
    \includegraphics[width=0.21\linewidth]{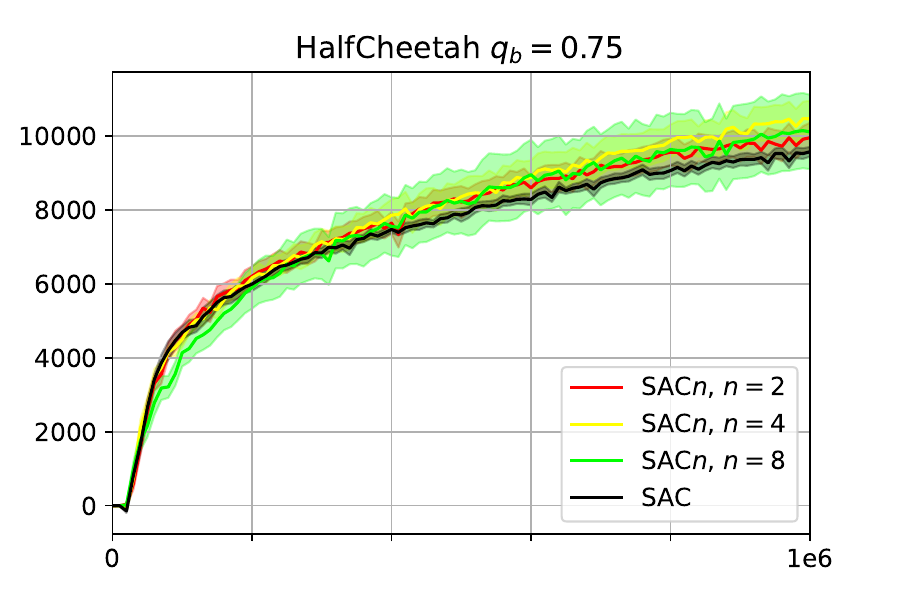}
    \hspace{-0.5cm}
    \includegraphics[width=0.21\linewidth]{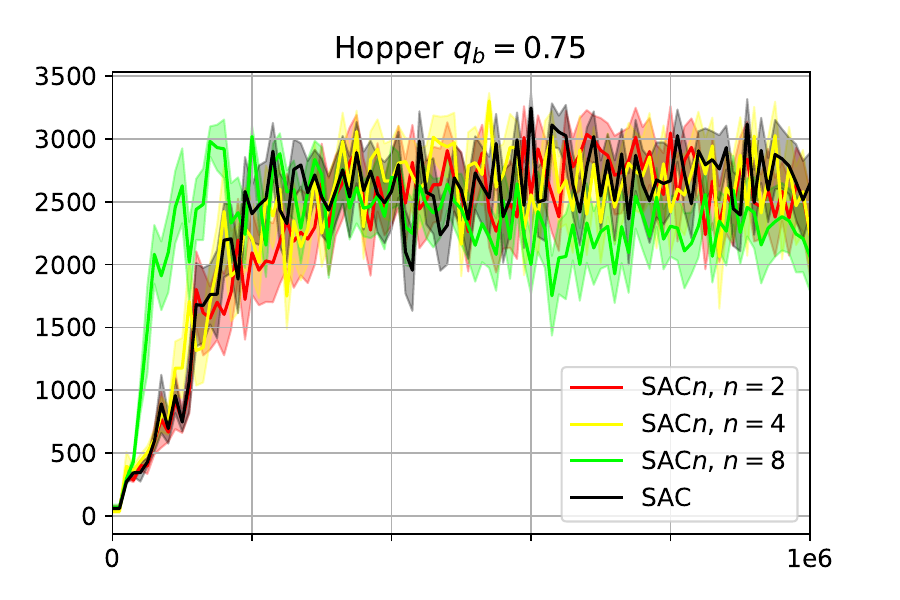}
    \hspace{-0.5cm}
    \includegraphics[width=0.21\linewidth]{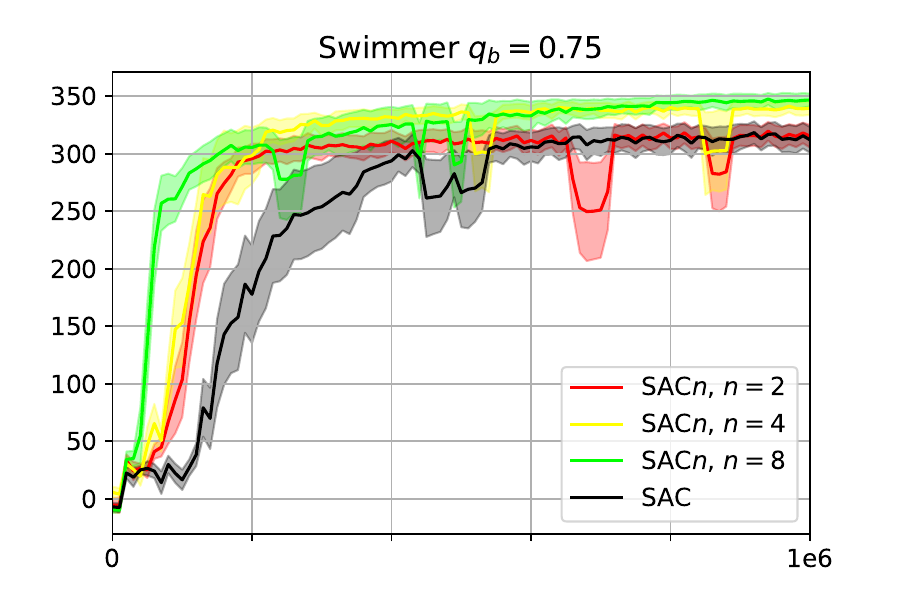}
    \hspace{-0.5cm}
    \includegraphics[width=0.21\linewidth]{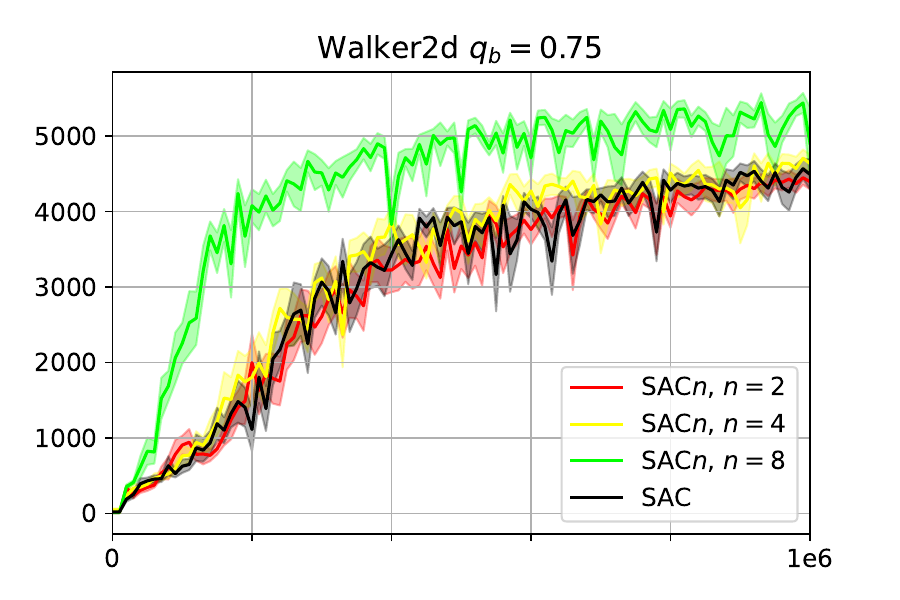}
    \hspace{-0.5cm}\\
    \includegraphics[width=0.21\linewidth]{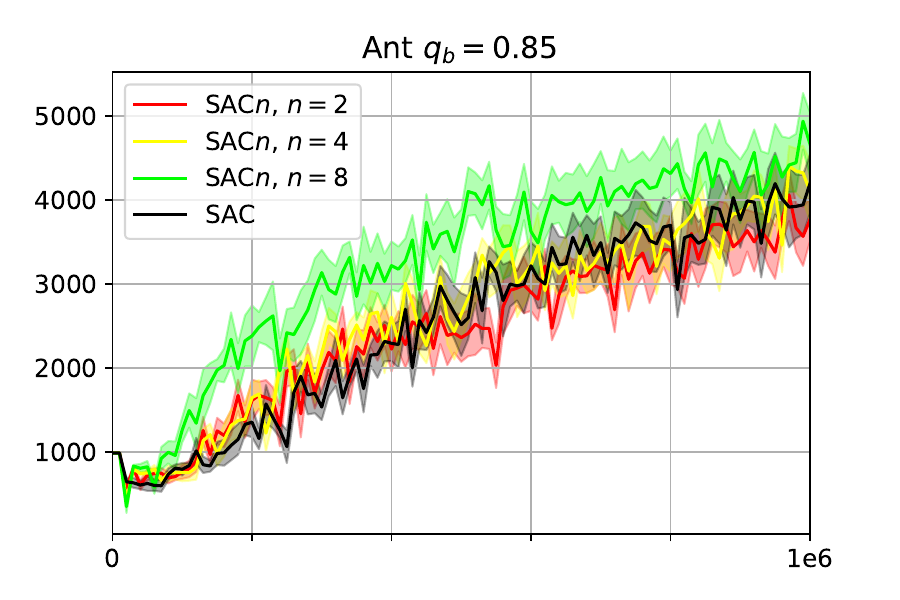}
    \hspace{-0.5cm}
    \includegraphics[width=0.21\linewidth]{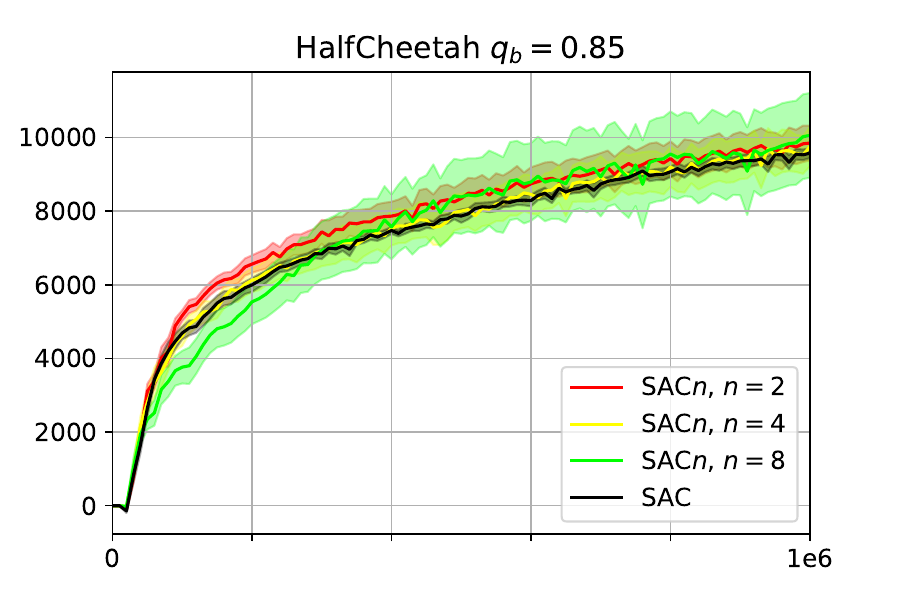}
    \hspace{-0.5cm}
    \includegraphics[width=0.21\linewidth]{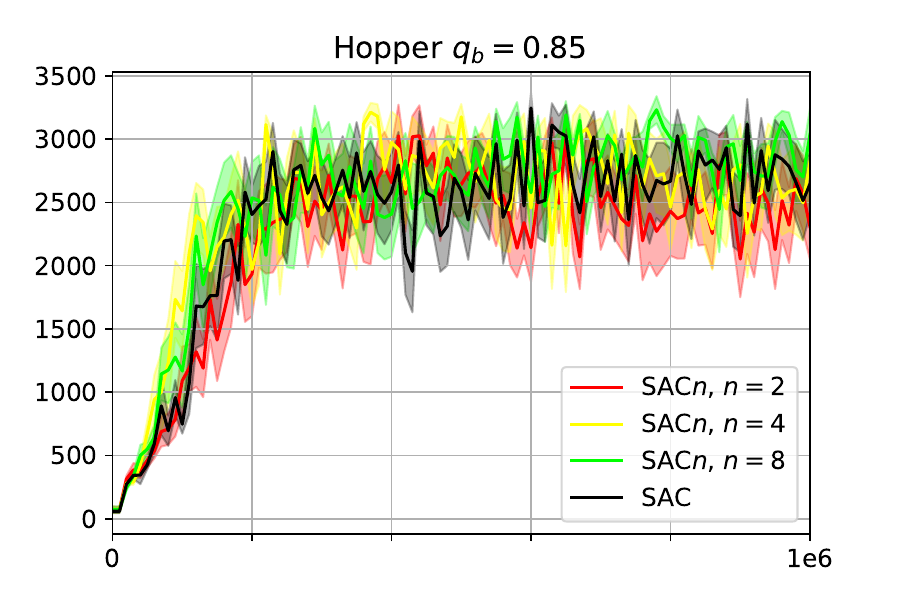}
    \hspace{-0.5cm}
    \includegraphics[width=0.21\linewidth]{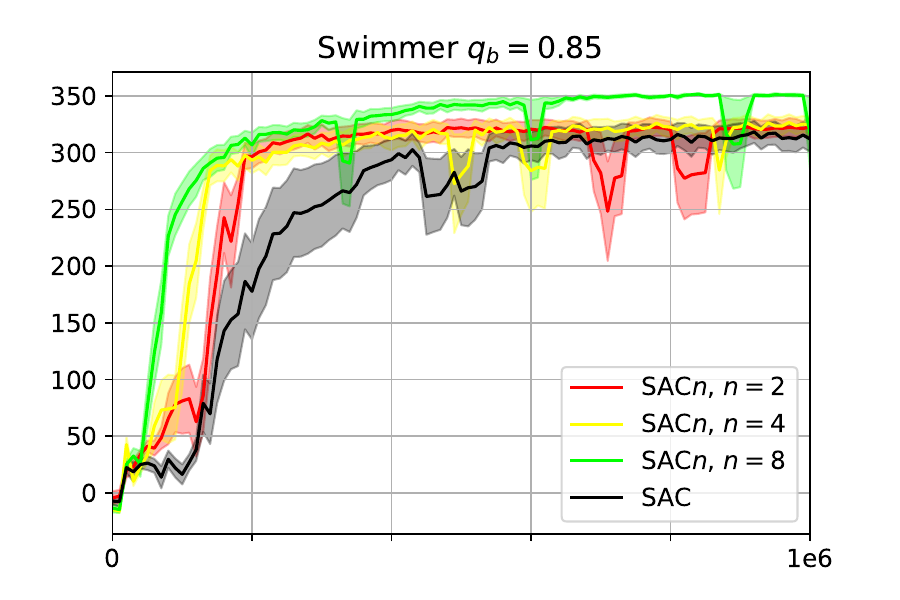}
    \hspace{-0.5cm}
    \includegraphics[width=0.21\linewidth]{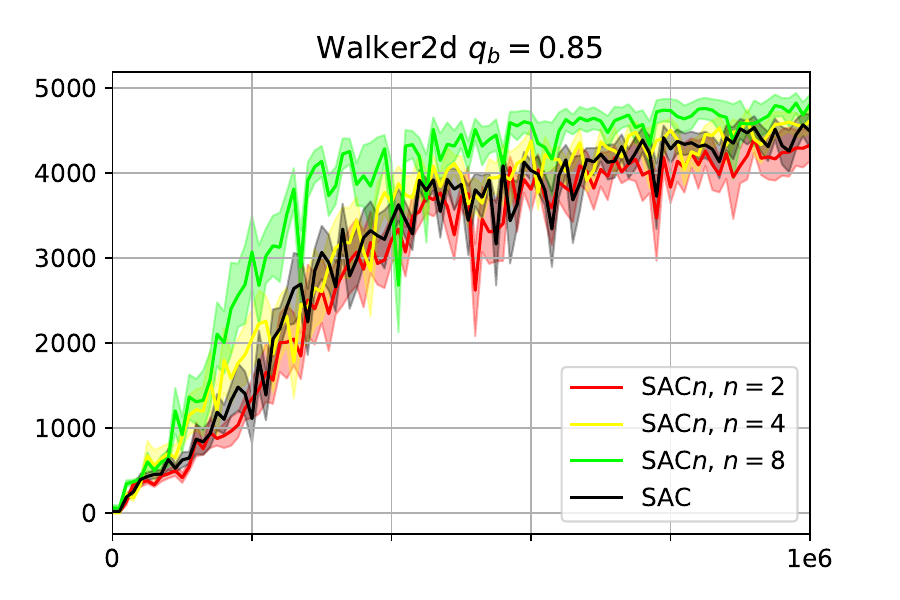}
    \hspace{-0.5cm}\\
    \includegraphics[width=0.21\linewidth]{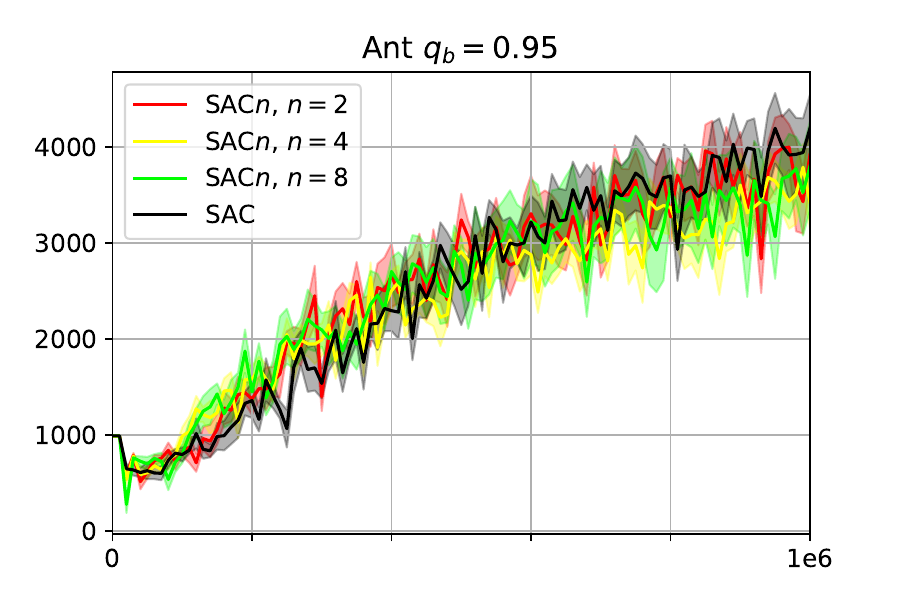}
    \hspace{-0.5cm}
    \includegraphics[width=0.21\linewidth]{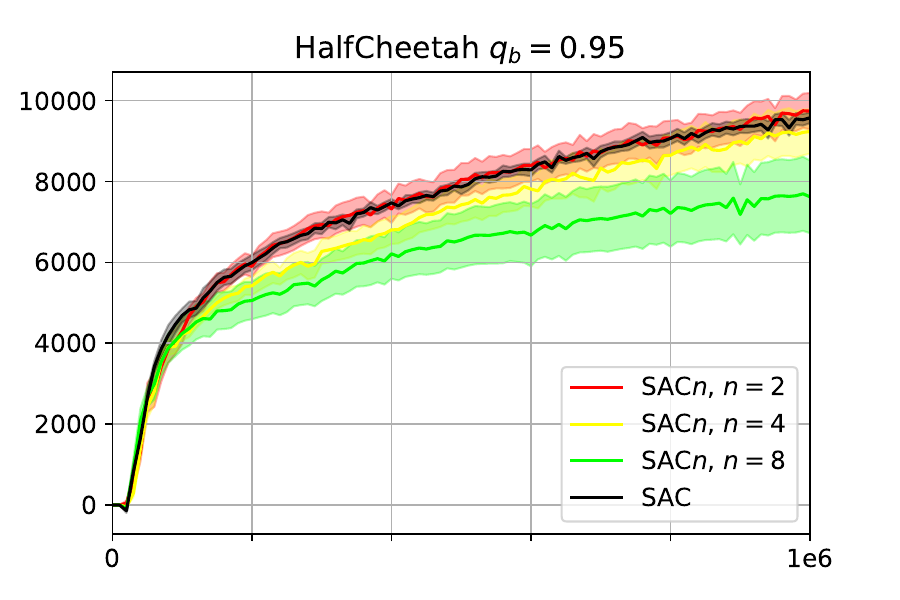}
    \hspace{-0.5cm}
    \includegraphics[width=0.21\linewidth]{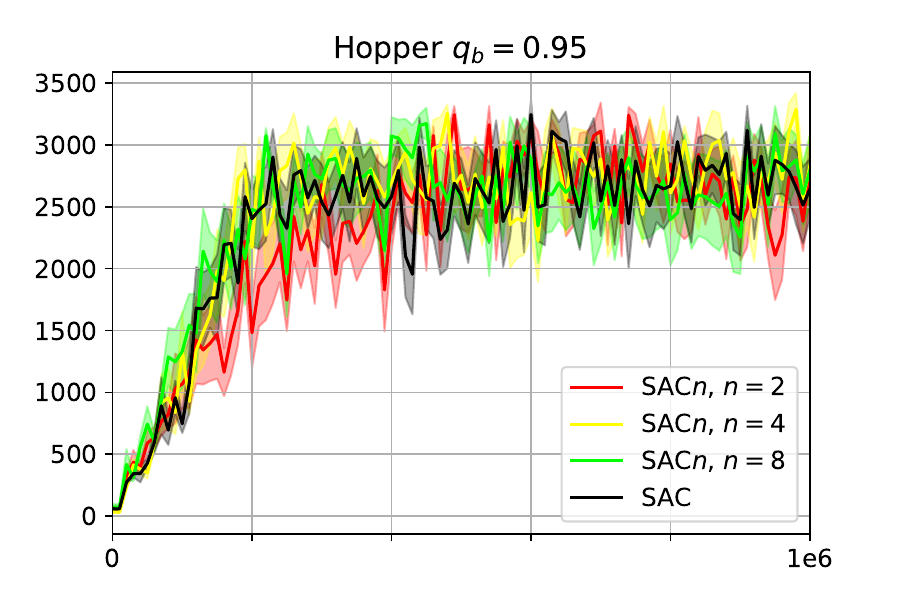}
    \hspace{-0.5cm}
    \includegraphics[width=0.21\linewidth]{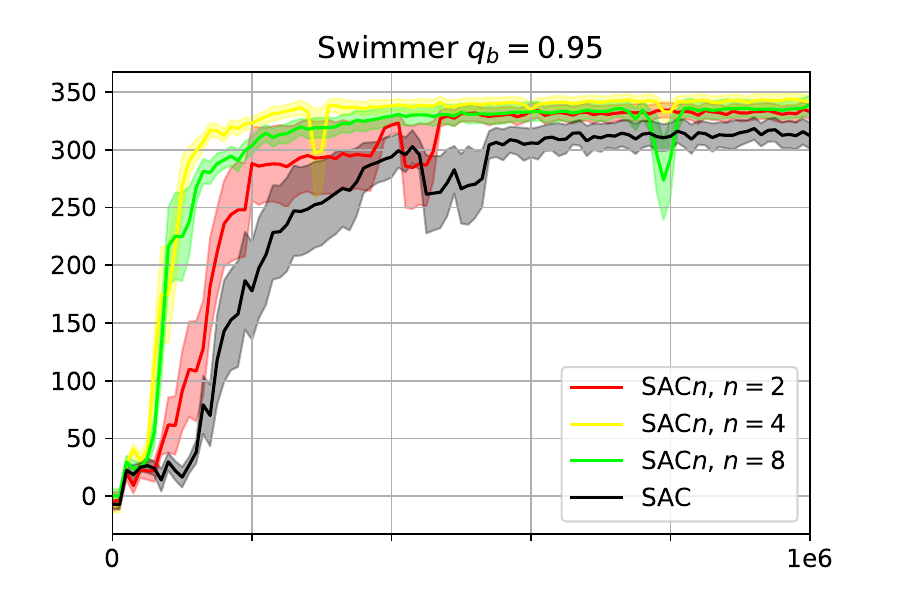}
    \hspace{-0.5cm}
    \includegraphics[width=0.21\linewidth]{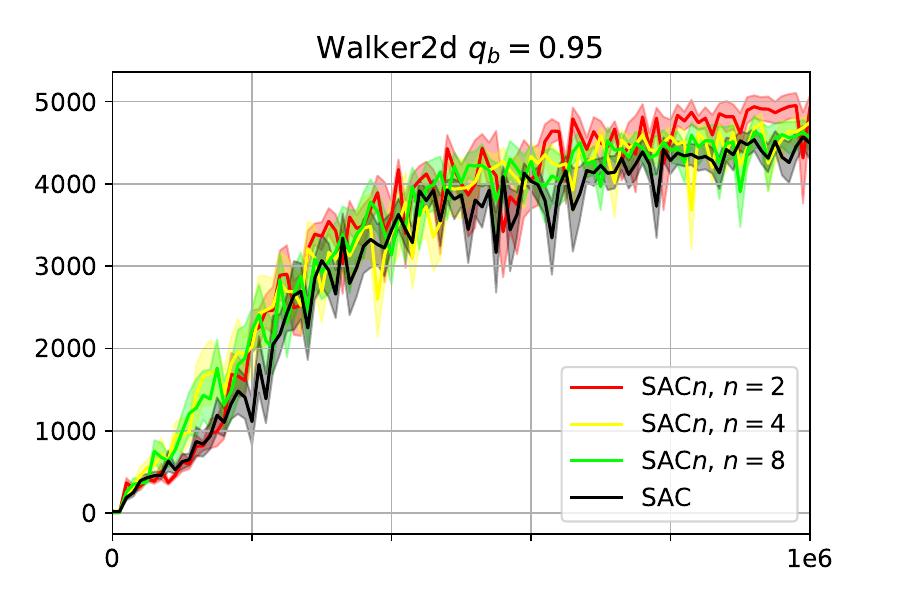}
    \hspace{-0.5cm}\\
    \caption{Learning curves for SAC$n$ with different values of $q_b$ hyperparameter.}
    \label{fig:b_ablation}
\end{figure*}

\begin{figure*}
    \centering
    \includegraphics[width=0.21\linewidth]{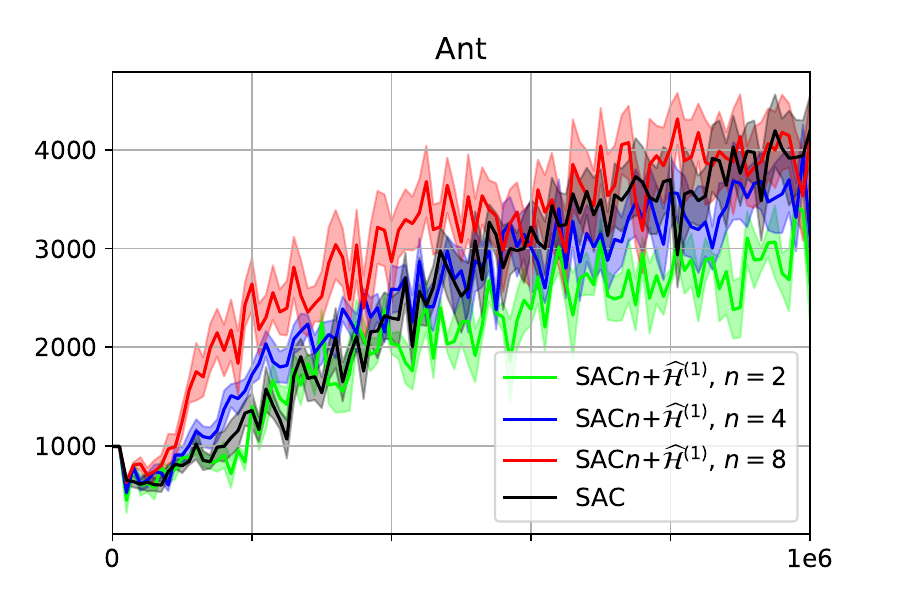}
    \hspace{-0.5cm}
    \includegraphics[width=0.21\linewidth]{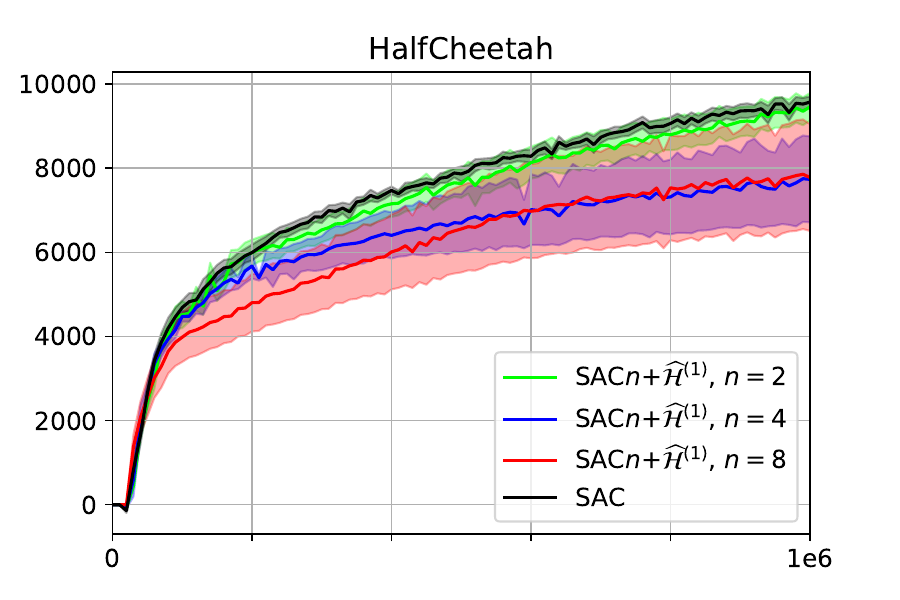}
    \hspace{-0.5cm}
    \includegraphics[width=0.21\linewidth]{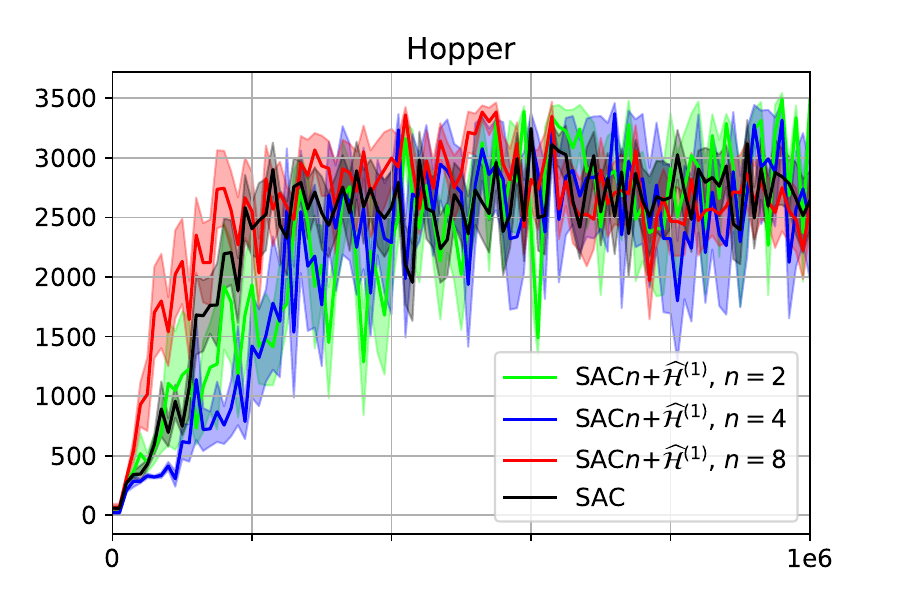}
    \hspace{-0.5cm}
    \includegraphics[width=0.21\linewidth]{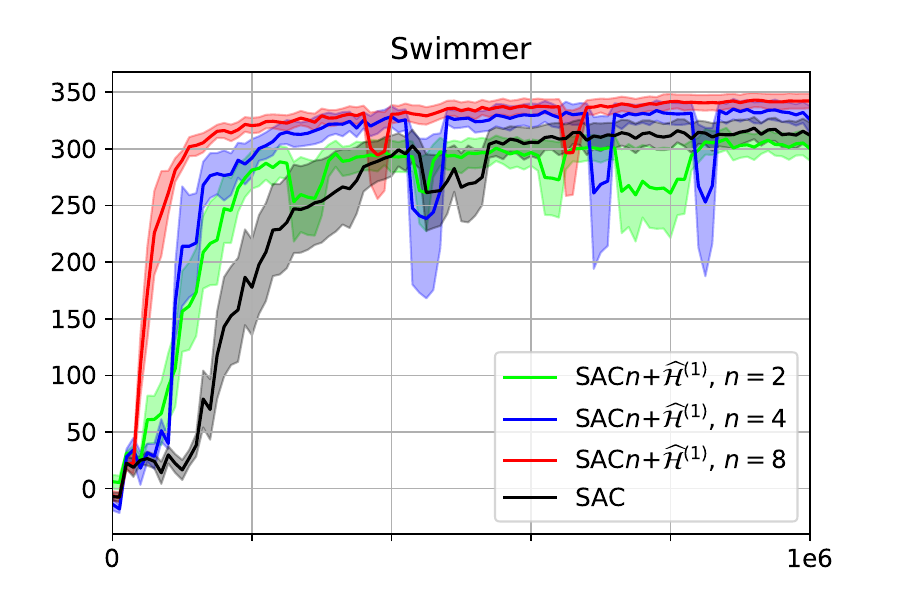}
    \hspace{-0.5cm}
    \includegraphics[width=0.21\linewidth]{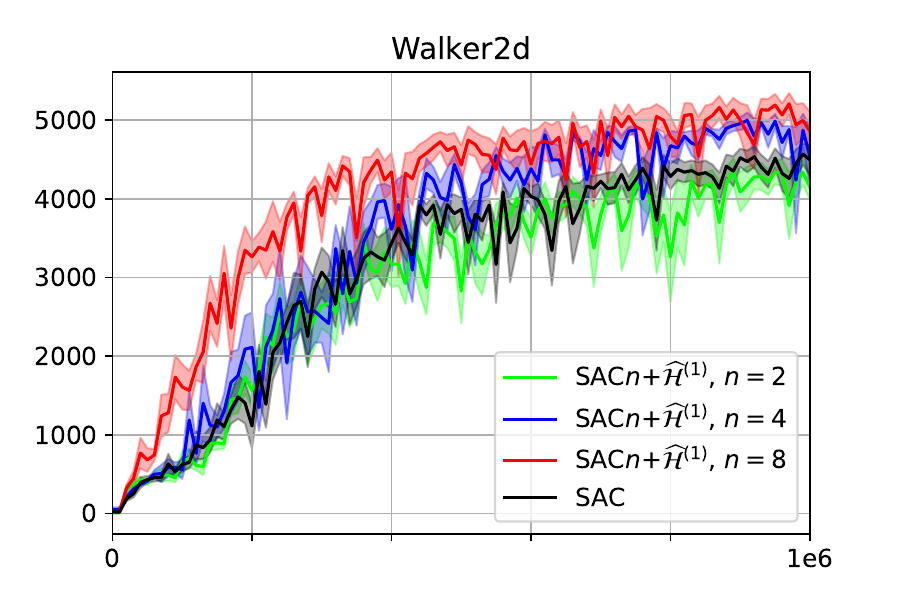}
    \hspace{-0.5cm}\\
    \caption{Learning curves for SAC$n$ without $\tau$-sampled entropy estimation.}
    \label{fig:no_sample_n_ablation}
\end{figure*}

\begin{figure*}
    \centering
    \includegraphics[width=0.21\linewidth]{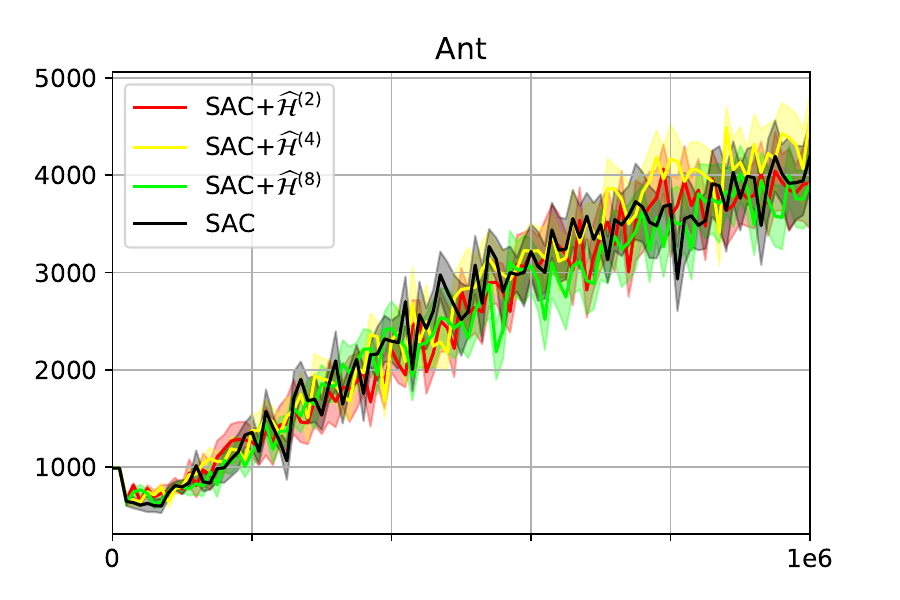}
    \hspace{-0.5cm}
    \includegraphics[width=0.21\linewidth]{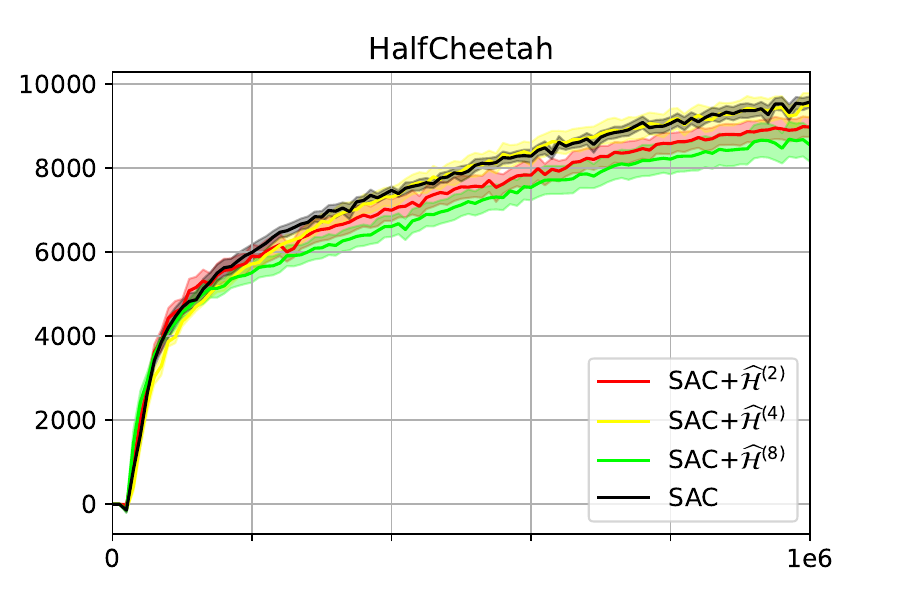}
    \hspace{-0.5cm}
    \includegraphics[width=0.21\linewidth]{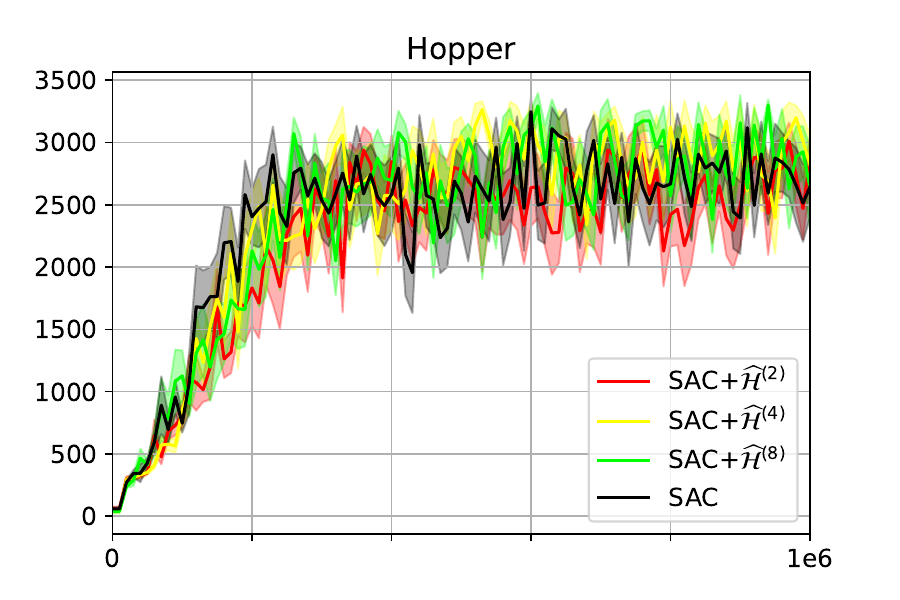}
    \hspace{-0.5cm}
    \includegraphics[width=0.21\linewidth]{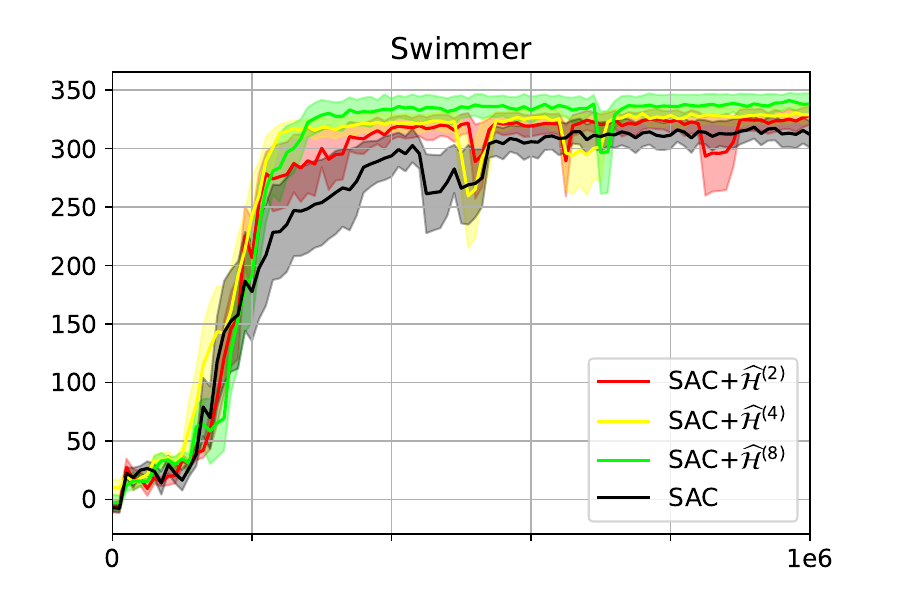}
    \hspace{-0.5cm}
    \includegraphics[width=0.21\linewidth]{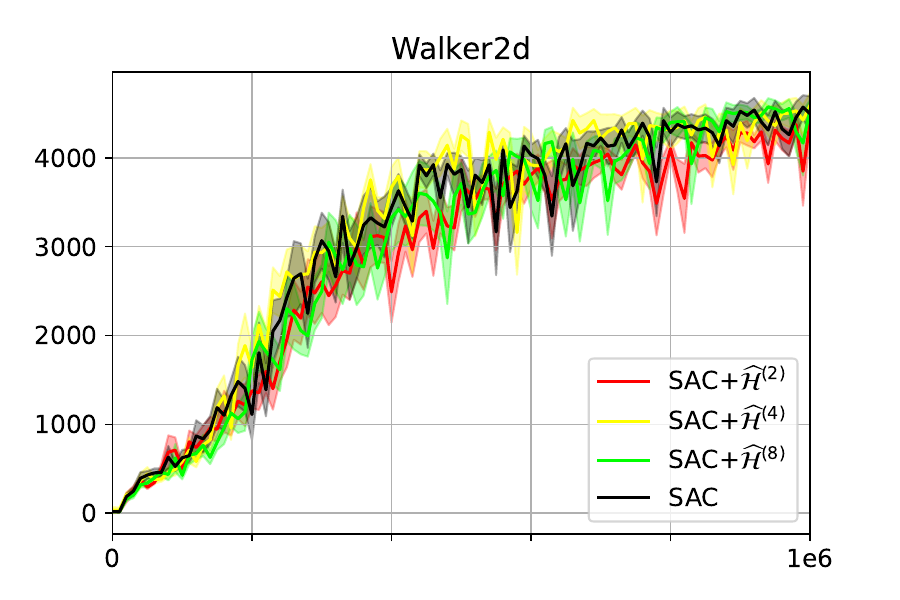}
    \hspace{-0.5cm}\\
    \caption{Learning curves for SAC with $\tau$-sampled entropy estimation.}
    \label{fig:sample_n_ablation}
\end{figure*}

Figures \ref{fig:b_ablation}, \ref{fig:no_sample_n_ablation}, and \ref{fig:sample_n_ablation} contain the learning curves for the experiments whose aim is to measure the influence of specific hyperparameters and components of SAC$n$ algorithm, presented in Subsection 5.4. Each result is averaged over last $3\cdot10^4$ time steps to account for large amplitudes of the learning curves for some environments and over 12 runs, and the results are presented together with their standard errors. Specifically, figure \ref{fig:b_ablation} contains the learning curves for different values of the $q_b$ and $n$ hyperparameters. Figure \ref{fig:no_sample_n_ablation} contains the learning curves for SAC$n$ without $tau$-sampled entropy estimation. Figure \ref{fig:sample_n_ablation} contains the learning curves for SAC with $tau$-sampled entropy estimation for different values of $\tau$.

\section{\uppercase{Hyperparameter settings}}\label{app:hyperparameters}

In this section, we provide values of all the used hyperparameters for the experiments reported in Section 5. For SAC, we used the values provided by \citet{raffin2020rlzoo3}, listed in the Table \ref{tab:hyperparameters}. For SAC$n$, we used the same values as for SAC where applicable. For SAC$n$, the default value of $q_b$ was $0.75$ and the values of $n$ were provided for each result.

\begin{table}
    \centering
    \caption{SAC and SAC$n$ hyperparameter values as provided for SAC by \citet{raffin2020rlzoo3}. $S$ - denotes the state space.}
    \label{tab:hyperparameters}
    \begin{tabular}{c|c}
        \hline
         Hyperparameter & Value  \\
         \hline
         Time steps & $10^6$ \\ 
         Batch size & $256$ \\ 
         Discount (Swimmer) & $0.999$  \\ 
         Discount (other) & $0.99$ \\
         Learning rate & $3\cdot10^{-4}$ \\ 
         Critic hidden layers & $\langle256,256\rangle$ \\ 
         Actor hidden layers & $\langle256,256\rangle$  \\ 
         Critic activation function & ReLU \\ 
         Actor activation function & ReLU \\ 
         Learning start & $10^4$ \\ 
         Training frequency & 1 \\
         Num. of training steps & 1 \\
         Target network update freq. & 1 \\
         Target network update coeff. & 0.005 \\
         Entropy target & $-dim(S)$ \\
         \hline
    \end{tabular}
\end{table}

\end{document}